\documentclass[10pt,onecolumn,letterpaper]{article}

\usepackage{times}
\usepackage{epsfig}
\usepackage{graphicx}
\usepackage{amsmath}
\usepackage{amssymb}
\usepackage{array}
\usepackage{adjustbox}
\usepackage{array}
\usepackage{cellspace}
\usepackage{hhline}

\graphicspath{{images/}}
\usepackage[utf8]{inputenc} 
\usepackage{multirow}
\usepackage{placeins}
\usepackage{lineno}
\usepackage{color}
\usepackage{hyperref}
\usepackage{subcaption}

\usepackage{amsmath}
\usepackage{amsfonts}
\usepackage{color}
\usepackage{textcomp}
\usepackage[table,xcdraw]{xcolor}
\pdfoutput=1
\graphicspath{{./figuras/results/}{./figuras/}}
\usepackage[table]{xcolor}

\newcommand{\MB}{\texttt{MB}}
\newcommand{\HRB}{\texttt{HRB}}

\newcommand{\ECB}{\texttt{ECB}}
\newcommand{\DCB}{\texttt{DCB}}
\newcommand{\IB}{\texttt{IB}}

\newcommand{\DBGESDPD}{\texttt{GESDPD}}
\newcommand{\DBKTP}{\texttt{KTP}}
\newcommand{\DBUNIHALL}{\texttt{UNIHALL}}

\newcommand{\DBEPFLLAB}{\texttt{EPFL-LAB}}
\newcommand{\DBEPFLCORRIDOR}{\texttt{EPFL-CORRIDOR}}
\newcommand{\DBGFPD}{\texttt{GFPD}}

\newcommand{\Fonescore}{F_{1score}}
\newcommand{\Precision}{Precision}
\newcommand{\PrecisionS}{P}
\newcommand{\Recall}{Recall}
\newcommand{\RecallS}{R}

\newcommand{\AlgDPDNet}{\texttt{DPDnet}}

\newcommand{\AlgDPDNetvtwo}{\texttt{PD3net}}

\newcommand{\AlgDPOM}{\texttt{DPOM}}
\newcommand{\AlgACF}{\texttt{ACF}}
\newcommand{\AlgYOLODepth}{\texttt{YOLO-Depth}}
\newcommand{\AlgYOLOsixteen}{\texttt{YOLO16}}

\newcommand{\AlgYOLOvthree}{\texttt{YOLO-V3}}
\newcommand{\AlgPCLMunaro}{\texttt{PCL-MUNARO}}
\newcommand{\AlgKINECTtwo}{\texttt{KINECT2}}
\newcommand{\AlgRCNN}{\texttt{RCNN}}
\newcommand{\AlgUNIHALL}{\texttt{UNIHALL}}
\newcommand{\AlgRGBCNN}{\texttt{RGBCNN}}
\newcommand{\AlgRGBCECDCNN}{\texttt{RGBCECDCNN}}

\newcommand{\AlgDPOMRef}{\AlgDPOM{}~\cite{Bagautdinov_CVPR_2015_dpom}}
\newcommand{\AlgACFRef}{\AlgACF{}~\cite{acf}}
\newcommand{\AlgYOLOvthreeRef}{\AlgYOLOvthree{}~\cite{yolov3}}
\newcommand{\AlgPCLMunaroRef}{\AlgPCLMunaro{}~\cite{munaro2014}}
\newcommand{\AlgKINECTtwoRef}{\AlgKINECTtwo{}~\cite{kinect2}}
\newcommand{\AlgRCNNRef}{\AlgRCNN{}~\cite{rcnn}}
\newcommand{\AlgRGBCNNRef}{\AlgRGBCNN{}~\cite{mva2017}}
\newcommand{\AlgRGBCECDCNNRef}{\AlgRGBCECDCNN{}~\cite{mva2017}}
\newcommand{\AlgUNIHALLRef}{\AlgUNIHALL{}~\cite{unihall}}

\definecolor{greenao}{rgb}{0.0, 0.5, 0.0}

\newcommand*{\mathcolor}{}
\def\mathcolor#1#{\mathcoloraux{#1}}
\newcommand*{\mathcoloraux}[3]{%
	\protect\leavevmode
	\begingroup
	\color#1{#2}#3%
	\endgroup
}

\definecolor{white}{rgb}{1,1,1}
\definecolor{orange}{rgb}{1,0.75,0}
\definecolor{111}{rgb}{0,1,0}
\definecolor{00}{rgb}{1.00000000000000000000,.49803921568627450980,0}
\definecolor{222}{rgb}{0.8,0.9568627451,0.56470588235}
\definecolor{111}{rgb}{0.8,0.9568627451,0.56470588235}

\begin{document}
\title{Towards Dense People Detection with Deep Learning and Depth images}

\author{David Fuentes-Jimenez, Cristina Losada-Gutierrez,\\
	Roberto Martin-Lopez,Javier Macias-Guarasa,\\
	 Daniel Pizarro, Carlos A.Luna\\
Universidad de Alcal\'a\\
{\tt\small \{d.fuentes, roberto.martin\}@edu.uah.es}\\
{\tt\small\{daniel.pizarro, cristina.losada,javier.maciasguarasa,carlos.luna\}@uah.es} 
\\
David Casillas-Perez\\
Universidad Rey Juan Carlos\\
{\tt\small \{david.casillas\}@urjc.es}}

\maketitle

\begin{abstract}
  This paper proposes a DNN-based system that detects multiple people from a single depth image. Our neural network
processes a depth image and outputs a likelihood map in image coordinates, where each detection
corresponds to a Gaussian-shaped local distribution, centered at the person's head. The likelihood map encodes both the number of detected
people and their 2D image positions, and can be used to recover the 3D position of each person using
the depth image and the camera calibration parameters. Our architecture is compact, using
separated convolutions to increase performance, and runs in real-time with low budget GPUs. We use
simulated data for initially training the network, followed by fine tuning with a relatively small amount of real data. 
We show this strategy to be effective, producing networks that generalize
to work with scenes different from those used during training. We thoroughly compare our method against
the existing state-of-the-art, including both classical and DNN-based solutions. Our method outperforms
existing methods and can accurately detect people in scenes with significant occlusions. 
\end{abstract}

\section{Introduction}
\label{sec:intro}

People detection and localization from cameras has received a great deal of attention from the scientific community recently, due to its multiple applications in different areas, such as security, video surveillance~\cite{villamizar2018,moro2018} or healthcare~\cite{lee2013context,wang2016human, Gavriilidis2018}. 
However, it remains an open problem, and presents several challenging tasks~\cite{zhang2016far,zhang2018}, especially in crowded scenes.

Early people detection methods used RGB images captured from a mainly
\emph{frontal} viewpoint. Methods such as ~\cite{Ramanan2007,
	TsongYi2010, ChiYoon2013, Slawomir2016, Aguilar2017} use traditional
computer vision algorithms, such as appearance models
in~\cite{Ramanan2007} or classic face detection
in~\cite{TsongYi2010}. These approaches obtained good results in
controlled spaces but struggle with the presence of partial occlusions,
motion blur and low resolution images.  

\begin{figure}[htbp]
	\centering
	\includegraphics[width=1.0\textwidth]{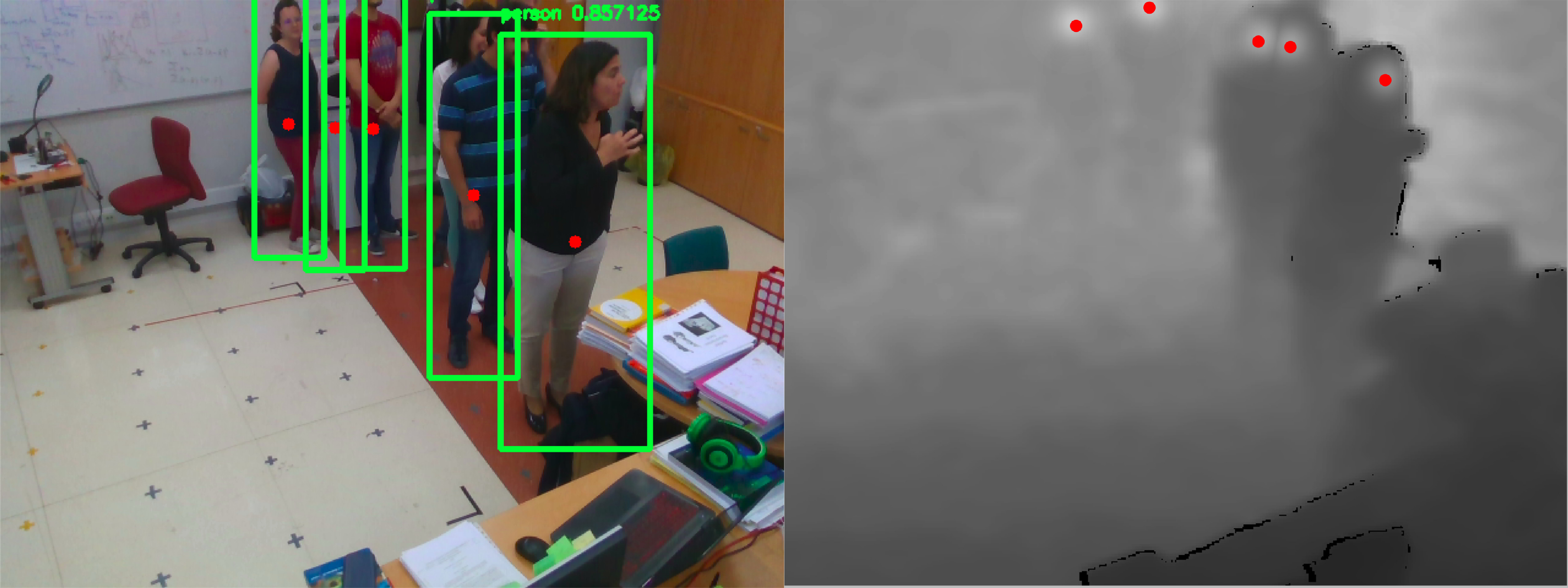}
	\caption{ Performance of our \AlgDPDNetvtwo{} vs the DNN-based method \AlgYOLOsixteen{}. On the left, \emph{a)} shows the people detection results obtainted by \AlgYOLOsixteen{}.  On the right, \emph{b} shows the results achieved by our \AlgDPDNetvtwo{}. Observe how \AlgYOLOsixteen{} presents false positives and true negatives. On the contrary, our system is able to correctly detect all people in that scene with strong oclussions.}
	\label{fig:introFig}
\end{figure}

Deep Neural Networks (DNN) have greatly improved the state-of-the-art in
several critical computer vision applications, such as object
detection~\cite{yolo2016}, semantic segmentation~\cite{cao2017},
classification~\cite{hoo2016} or activity
recognition~\cite{Hayashi15,zhang2015human}. Similarly, DNN-based
people detection methods using RGB
data~\cite{bochkovskiy2020yolov4,du2019spinenet,8954436} have
considerably improved over the classical algorithms. However, DNN-based
methods also have significant drawbacks, such as the large amount of
labeled data needed for training and the requirement of dedicated
processing units to run and train the network. Besides, recent DNN-based
methods still present low accuracy in highly cluttered scenes (see
Fig.~\ref{fig:introFig}).  

People detection using depth images is a less popular topic in the
literature, mainly because depth cameras are not as widely available as
RGB cameras. Nonetheless, using depth images has significant advantages:
\textit{1)} Depth images naturally disambiguate objects at different
depths, which helps to process occlusions in crowded scenes. \textit{2)}
Depth information is less complex than RGB information as it is not
affected by appearance or light changes. \textit{3)} Once detected in
the image, positions of people in 3D are directly available using depth
information, which is a desirable feature in many applications. 

There exist several depth-based people detection methods in the
literature, and some of them also include RGB information. Recent
DNN-based approaches obtain the best detection results in this
category. However, they bear important limitations. Some methods are
specific to \textit{zenithal}
viewpoints~\cite{delpizzo2016counting,luna2017,dpdnet,zhou2017}, which
reduces occlusions and makes the problem less ambiguous, but limits the
field-of-view. Others use a conventional \textit{frontal}
viewpoint~\cite{rcnn,zhou2017}, which covers a wider range of
applications. \cite{rcnn,zhou2017} use a region proposal method to
detect candidates, and a classifier to select the positive regions that
correspond to a person. This strategy is not optimal, especially for
densely populated scenes, and it is not efficient, as its complexity
depends on the number of possible candidates detected in the image.   

This paper proposes a new DNN-based approach, that we call
\AlgDPDNetvtwo{}, for detecting multiple people from a single depth
image acquired using a camera in an elevated frontal position, see
Fig.~\ref{fig:introFig}. Our method has the following contributions: 
1) Our network architecture is fully convolutional and very efficient by using spatially separable convolutions. It runs in real-time with low-cost GPU and CPU architectures.
2) We train the neural network end-to-end with synthetically generated depth images and then we fine-tune the network with a small number of annotated real images.  
The paper shows that this strategy leads to accurate and generalistic detectors that work well in general scenes.    
3) Our method recovers a dense likelihood map that effectively detects multiple people in crowded scenes (see Fig.~\ref{fig:introFig}).   
4) We outperform the existing state-of-the-art, including both classical and DNN-based approaches.  
5) The proposed method works with different cameras and depth sensing technologies. 
6) Our method does not have a maximum restriction of detections per image. 

The rest of the paper is structured as follows: Section~\ref{sec:sota}
reviews the latest state-of-the-art methods focused on the people
detection field.  Section~\ref{sec:tofnet} explains in detail our
DNN-based proposal, describing its architecture and the training
procedure.  Section~\ref{sec:experimental-work} shows the experimental
setup developed to evaluate our approach. This section includes a
thorough comparison with the main state-of-the-art methods over a wide
range of publicly available datasets.  Finally,
Section~\ref{sec:conclusions} describes the main conclusions and propose
some future lines.

\section{Previous works}
\label{sec:sota}

People detection methods are classified in this section according to
three main criteria: \emph{a)} the type of information used, \emph{b)}
the type of algorithm used, distinguishing between classical and
DNN-based strategies, and finally \emph{c)} the camera's point of view.
Table~\ref{tab:prevWorks} shows the main state-of-the-art people
detection methods and their corresponding classification.

\begin{table}[htpb]
	\begin{center}
		\caption{Classification of the main people detector methods in the state-of-the-art.}
		\label{tab:prevWorks}
		\resizebox{0.91\textwidth}{!}{
			\begin{tabular}{|c|c|c|c|c|}
				\hline
				Reference                                                   & Algorithm & Input Inform. & Camera viewpoint & Description \\ \hline\hline
				\cite{Ramanan2007}                         & Classical & RGB               & Frontal              &   Tracking by model-building and detection          \\ \hline
				\cite{Chan2008}\footnotemark[1]            & Classical & RGB               & Frontal              & Privacy-preserving system based on a mixture of dynamic textures motion model                     \\ \hline
				\cite{TsongYi2010}                         & Classical & RGB               & Frontal              &   People counting system based on face detection            \\ \hline
				\cite{ByoungKyu2012}                       & Classical & RGB-D             & Zenithal             &  RGB and Depth fusion for people detection          \\ \hline
				\cite{Zhang2012}\footnotemark[1]           & Classical & Depth             & Zenithal             & Unsupervised people counting via vertical Kinect Sensor                      \\ \hline
				\cite{Stahlschmidt2013,Stahlschmidt2014}\footnotemark[1]   & Classical & Depth & Zenithal         & Differences from the ground plane are used to develop regions of interest                      \\ \hline
				\cite{ChiYoon2013}                         & Classical & RGB               & Frontal              &   People counting based in statical and moving detection points.                     \\ \hline
				\cite{zhu2013human}                        & Classical & RGBD              & Zenithal             & Adaboost algorithm built from weak classifiers for detecting people                       \\ \hline
				\cite{galvcik2013real}                     & Classical & Depth             & Frontal              & Real-Time People Detector with minimum-weighted bipartite graph matching                    \\ \hline
				\cite{du2015}                              & DNN       & RGB               & Zenithal             & Optimization of pedestrian detection with semantic tasks                      \\ \hline
				\cite{liu2015}                             & Classical & RGB-D             & Zenithal             &  People detection taking different poses in cluttered and dynamic environments                      \\ \hline
				\cite{delpizzo2016counting}                & Classical & RGB-D             & Zenithal             &  Depth-RGB and both people detector with crossing-path points          \\ \hline
				\cite{vera2016counting}                    & DNN       & Audio             & -                    & Cooperating network for people detection                      \\ \hline
				\cite{Slawomir2016}                        & Classical & RGB               & Frontal              &   Brownian covariance descriptor                     \\ \hline
				\cite{Aguilar2017}                         & Classical & RGB               & Frontal              &   Cascade classifier with salience map for pedestrian detection                     \\ \hline
				\cite{wang2017}                            & DNN       & RGB               & Zenithal             &   Multi-layer regional-based convolutional for crowded scenes                     \\ \hline
				\cite{zhao2017}                            & DNN       & RGB               & Frontal              &  Real-time detection based on physical radius-depth detector                   \\ \hline
				\cite{luna2017}                            & Classical & Depth             & Zenithal             &  ToF people detector based on depth information                      \\ \hline
				\cite{ren2017}                             & Classical & RGB-D             & Zenithal             &  Parallel deep feature extraction from RGB and Depth simultaneously                      \\ \hline
				\cite{zhou2017}                            & Classical & RGB-D             & Zenithal             &  Depth-encoding scheme which enhances the information for classification                      \\ \hline
				\cite{hu2018}                              & Classical & RGB-D             & Zenithal             &  Uses the 3D Mean-Shift with depth constraints to multi-person detection     \\ \hline
				\cite{Verma2019}                           & Classical & RGB               & Zenithal             &  Fuzzy-based detector based on CCA components       \\ \hline
				\cite{dpdnet}\footnotemark[1]              & DNN       & Depth             & Zenithal             &  Encoder-decoder DNN blocks with refinement.                      \\ \hline
			\end{tabular}
		}
	\end{center}
\end{table}
\footnotetext[1]{There exists privacy requirements during the people detection task. Identification of people in scene is forbidden.}

Classical approaches for people
detection~\cite{Ramanan2007,TsongYi2010,ChiYoon2013,Slawomir2016,Aguilar2017}
use conventional RGB images as input with \emph{frontal} camera
poses. Within these approaches, \cite{Ramanan2007} proposes
a method based on appearance models, whereas \cite{ChiYoon2013} suggests
an approach for people detection using interest point
classification. Other alternatives for RGB people detection are based in
face detection \cite{TsongYi2010}, image descriptors based on Brownian
motion statistics \cite{Slawomir2016} or HAAR-LBP and HOG cascade
classifiers combined with Saliency Maps~\cite{Aguilar2017}.

The recent technology improvements and the availability of large
annotated image datasets~\cite{imagenet2012} have allowed Deep Neural
Networks based techniques to be widely used in computer vision task such
as object detection~\cite{yolo2016}, semantic
segmentation~\cite{cao2017} or classification~\cite{hoo2016}.
Regarding the people detection task, we can find several approaches
based on DNNs. Works like \cite{wang2017,zhao2017} use DNNs as feature
extractors, whereas \cite{du2015} proposed a novel DNN model that jointly
carry out people detection with semantic tasks.

As previously mentioned, one of the classification criteria used is the
location and pose of the camera. The proposals that position the camera
with a frontal viewpoint, works fine but they have certain
drawbacks. This location is highly sensitive to occlusions, and
consequently, the people detection performance degrades. In order to
solve this problem, some works proposed the use of certain alternatives
and constraints. One typical approach is changing the camera location to
a top view configuration, also referred to as zenithal or overhead
viewpoint in the literature. This is the case of \cite{Verma2019} which
proposes a fuzzy-based people detector from RGB data and a zenithal
location of the camera. Other works proposed the use of depth cameras,
that are considerably more insensitive to oclusions that conventional
RGB cameras. Some works such as \cite{luna2017} exclusively uses depth
information from an overhead camera, while others evaluates both RGB and
depth information from a top view camera
\cite{ByoungKyu2012,delpizzo2016counting}. Finally, some authors
propose the fusion of RGB and depth data (RGB-D) to adress the
oclussions effect~\cite{liu2015,zhou2017,ren2017,hu2018}.

The overhead viewpoint greatly solves the occlusions problem, but bring
some drawbacks, such as the reduction of the field of view, as the
covered area is directly related to the camera location height which is
limited for indoor applications. In addition to this, the use of
non-overhead perspectives is necessary in many real applications,
especially in video-surveillance. For all these reasons, out proposal in
this paper adopts a slightly elevated frontal camera pose, despite its
sensitivity to occlusions. The entry of DNN-based methods has mitigated
the effects of occlusions in this type of systems, mainly due to their
high learning capacity and robustness to occlusions, improving the
detection results achieved by classical
algorithms~\cite{wang2017,du2015,zhao2017,dpdnet}. 


Once the primary requirements of the people detection task were met,
other problems began to be considered. One of the main ones is related
to preserving privacy, especially in applications with restrictive
privacy policies in public spaces. This problem is more accentuated if
we deal with systems that use RGB information, so that some works like
\cite{Chan2008} proposed the use of remote cameras or low camera
resolutions. The main consequence of these proposals is the accuracy
reduction and the increased effect of occlusions in performance. Due to
the problems associated with RGB systems, alternatives began to be
studied in depth-only based methods, which do not easily allow the
recognition of people's identity. As a result, approaches based on depth, mostly with
cameras in overhead position, appeared to work well. Among the
depth-based methods, the proposal by~\cite{Zhang2012} is based on a
maximum detector followed by a water-filling algorithm,
while~\cite{Stahlschmidt2013,Stahlschmidt2014} filter depth images using
the normalized Mexican Hat Wavelet. Both proposals reduce their
detection rate when people are very close to each other, or cross their
paths. Besides, false positives appear if there are body parts different
from the head, such as hands, closer to the camera. To address these
drawbacks, several proposals include a classification stage to
discriminate people from other elements in the
scene~\cite{galvcik2013real,vera2016counting,zhu2013human,luna2017},
thus reducing these false positives. Again, al these proposals using a
camera in a top-view configuration, significantly reduce the field of
view.

In our DNN-based people detection proposal, we work with depth maps
taken from a slightly elevated front view camera position, thus
addressing the privacy preservation problem, while obtaining high accuracy
and solving to some extent the occlusions problem.

\section{\AlgDPDNetvtwo{} People Detector}
\label{sec:tofnet}

\subsection{Problem Formulation}
\label{sec:problem_formulation}

This paper propose \AlgDPDNetvtwo{}, a one shot CNN-based multiple people detector in depth maps. This system inherits part of its structure from \AlgDPDNet{}~\cite{dpdnet}, that solves people detection in overhead images. The proposed system and \cite{dpdnet} differ each other in both technically and in functionallity terms.

In functionality terms, \AlgDPDNetvtwo{} generalizes the previous network to non-overhead camera poses. Non-overhead camera poses are frequent in practical scenarios, much more than the restricted overhead view.

\begin{figure}[!htbp]
	\centering
	\includegraphics[width=0.7\columnwidth]{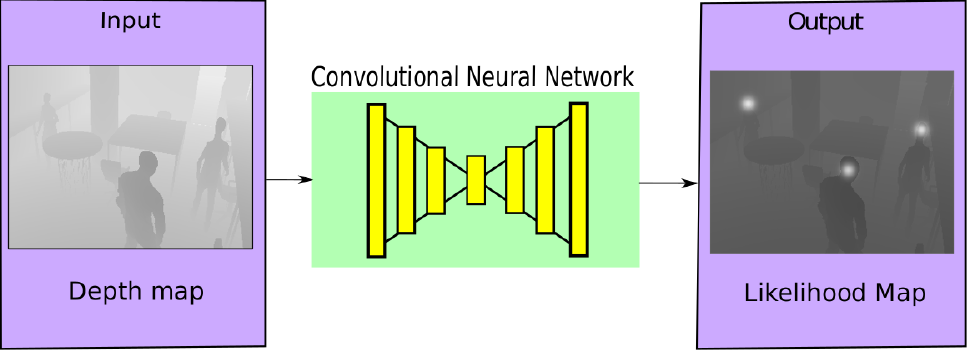}
	\caption{Depth input image and likelihood output map.}
	\label{fig:inputOutput1}.
\end{figure}

The non-overhead camera location is usually better than the overhead one in what respect to the greater amount of information it provides about the environment. In addition, it eliminates the need of using a specific camera pose, since many current surveillance systems cannot assume that restriction.  \AlgDPDNetvtwo{} receives a depth input map as input
and produces two different outputs, see Figure~\ref{fig:inputOutput1}. The CNN proposed here provides
an output represented by a likelihood map that has the same size as the input image, i.e. it is spatially coincident. This output map shows the detections of people made by the system in pixel coordinates as we can see in the Figure~\ref{fig:inputOutput1}.

Technically, the \AlgDPDNetvtwo{} network core inherits the
Resnet50~\cite{resnet16} internal blocks structure. This structure was also used by \AlgDPDNet{} but now it is meticulously revisited to fulfill the demanding real time requirements. At the neuron level, \AlgDPDNetvtwo{} adds the \emph{Leaky ReLU} activation function to the neuron, which clearly increment the speed of the gradient computation during the backpropagation and solves the dying ReLU problem, still remaining its non-linearity. At the layer level, \AlgDPDNetvtwo{} introduces the \emph{spatially separable convolutions} which considerably reduces the number of parameters and, consequently, the number of computations.  

Finally, at structural level, a thorougly study of the number of layers was developed removing all layers whose detection improvements were negligible and slows down the people detection. Identity, deconvolutional and convolutional blocks were refined to minimize its number of parameters. Putting all these improvements together, \AlgDPDNetvtwo{} achieves the sought real time with the best people detection rates, the most important contribution of the paper.


\AlgDPDNetvtwo{} use the output format proposed by~\cite{dpdnet}, in which we used a likelihood map with 2D Gaussian distributions in the center of the detected persons. This format allows using the center of the distributions as the 2D positions of the persons. In contrast to the overhead problem solved in~\cite{dpdnet}, when moving to an elevated frontal position, the occlusions are one of the main problems that we
have to face. The solution we propose to deal with them is the use of variable 2D Gaussian distributions which adapt themselves to avoid their overlap, allowing the system to become more robust against hard-occlusions.

The proposed method has certain pros and cons. In the case of the
former, this system allows the detection of multiple people without ambiguity and in addition, the processing time is completely independent of the number of users detected. In the meantime, by contrast, firstly this system depends quite a lot on the hardware used, especially in terms of the quality of the 3D information obtained and secondly in terms of the positioning of the camera, since being a non-restrictive positioning requires the system to be much more general in its operation
and to have a high level of independence from the camera pose.

\subsection{Architecture of the \AlgDPDNetvtwo{} Network}
\label{sec:Arquitecture}

\begin{figure*}[!htbp]
	\centering
	\includegraphics[width=0.8\textwidth]{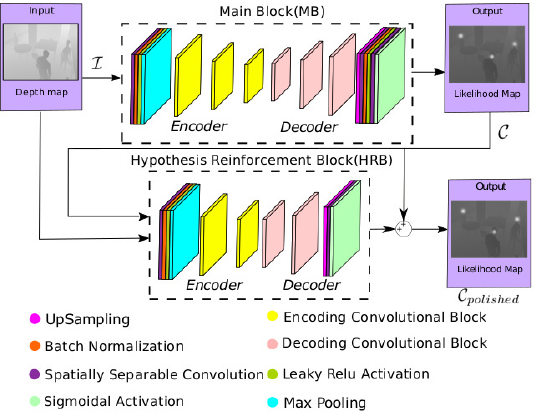}
	\caption{System's Architecture.}
	\label{fig:BP_BR}
\end{figure*}

Here we proposed a new upgraded architecture, shown in
Figure~\ref{fig:BP_BR}, for \AlgDPDNetvtwo{}, where its basis its
inherited from its predecessor~\cite{dpdnet}.

The base structure has similarities with the~\cite{dpdnet}
architecture. It is composed of two important blocks that are \emph{the main block (\MB{})} and the \emph{the hypothesis reinforcement block (\HRB{})}. These blocks are modeled by encoder-decoder architectures, which are traditionally used in tasks such as semantic segmentation~\cite{segnet,erfnet}, 3D reconstruction, registration~\cite{deepsft,hdm_net}, and deep fakes detection~\cite{Deepfake,Deepfake2}.

The input depth image $\mathcal{I}$ with size $240\times 320$ is
processed by the \MB{}, which generates the first likelihood map
$\mathcal{C}$ (also with size $240\times 320$), which is then
concatenated with the input image $\mathcal{I}$, so that the input tensor of the \HRB{} $\mathcal{I}_{2}$ is composed by the concatenation of $\mathcal{I}$ and $\mathcal{C}$, and has a size of $240\times 320\times 2$.  The \HRB{} polishes the initial likelihood map $\mathcal{C}$ and uses it as an output initialization, obtaining the final refined likelihood map $\mathcal{C}_{polished}$ ($240\times 320$), that validates, corrects and refines the predictions of the first map $\mathcal{C}$, thus improving its final results in
$\mathcal{C}_{polished}$. The detailed structure of
the \MB{} can be seen in Table~\ref{tab:mainBlock}.

\begin{figure}[!htbp]
	\begin{center}
		\begin{subfigure}[t]{0.49\textwidth}
			\includegraphics[width=\textwidth]{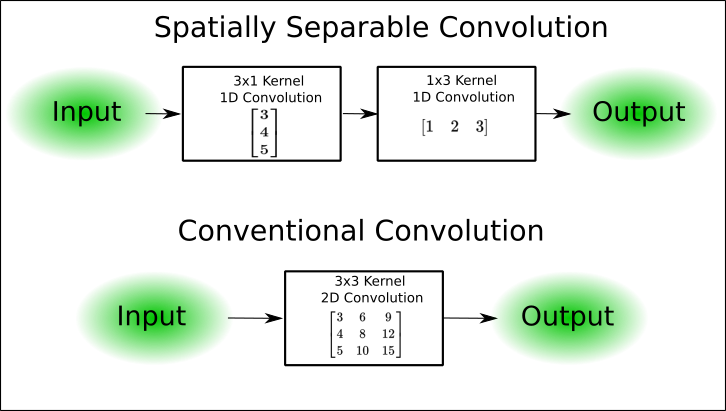}
			\caption{Example comparison.}
			\label{fig:spatial_conv}
		\end{subfigure}
		\begin{subfigure}[t]{0.49\textwidth}
			\includegraphics[width=\textwidth]{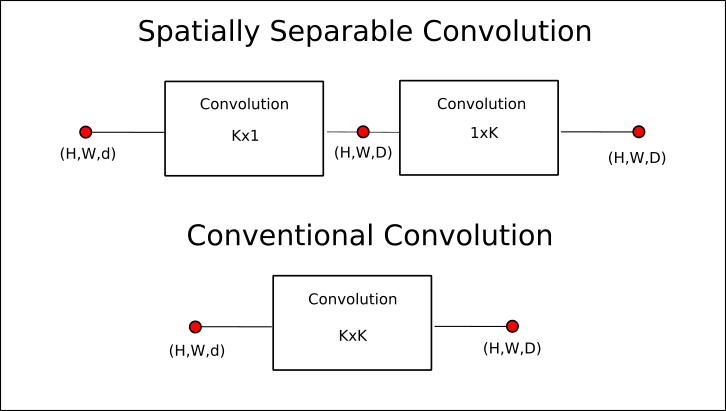}
			\caption{General formulation}
			\label{fig:conv_spatial}
		\end{subfigure}
	\end{center}
	\caption{Differences between conventional and spatially separable convolutions.}
	\label{fig:conv_spatial_all}
\end{figure}

\begin{table}[!htbp]
	\centering
	\caption{Detailed architecture of the main block.}
	\label{tab:mainBlock}
	\resizebox{0.5\columnwidth}{!}{%
		\begin{tabular}{| l | c | c | }
			\hline
			\multicolumn{3}{|c|}{\textbf{Main Block (\MB{})}} \\
			\hline
			\hline
			\multicolumn{1}{|c|}{\textbf{Layer}} & \textbf{Output size} & \textbf{Parameters} \\
			\hline
			\hline
			Input & $240\times320\times1$ & - \\
			\hline
			Convolution & $120\times160\times64$ & kernel=(7, 7) / strides=(2, 2) \\
			\hline
			BN & \multicolumn{2}{|c|}{-} \\
			\hline
			Activation & \multicolumn{2}{|c|}{Leaky ReLU} \\
			\hline
			Max Pooling & $40\times53\times64$ & size=(3, 3) \\
			\hline
			\hline
			\ECB{} & $40\times53\times256$ & \begin{tabular}{@{}c@{}}kernel=(3, 3) / strides=(1, 1) \\ (a=64, b=64, c=256) \end{tabular} \\
			\hline
			\ECB{} & $20\times27\times512$ & \begin{tabular}{@{}c@{}}kernel=(3, 3) / strides=(2, 2) \\ (a=128, b=128, c=512)  \end{tabular} \\
			\hline
			\ECB{} & $10\times14\times1024$ & \begin{tabular}{@{}c@{}}kernel=(3, 3) / strides=(2, 2) \\ (a=256, b=256, c=1024)  \end{tabular} \\
			\hline
			\hline
			\DCB{} & $10\times14\times256$ & \begin{tabular}{@{}c@{}}kernel=(3, 3) / strides=(1, 1) \\ (a=1024, b=1024, c=256)  \end{tabular} \\
			\hline
			\DCB{} & $20\times28\times128$ & \begin{tabular}{@{}c@{}}kernel=(3, 3) / strides=(2, 2) \\ (a=512, b=512, c=128)  \end{tabular} \\
			\hline
			\DCB{} & $40\times56\times64$ & \begin{tabular}{@{}c@{}}kernel=(3, 3) / strides=(2, 2) \\ (a=256, b=256, c=64)  \end{tabular} \\
			\hline
			\hline
			Cropping & $40\times54\times64$ & cropping=[(0, 0) (1, 1)] \\
			\hline
			Up Sampling & $120\times162\times64$ & size=(3, 3) \\
			\hline
			Convolution & $240\times324\times64$ & kernel=(7, 7) / strides=(2, 2) \\
			\hline
			Cropping & $240\times320\times64$ & cropping=[(0, 0) (2, 2)] \\
			\hline
			BN & \multicolumn{2}{|c|}{-} \\
			\hline
			Activation & \multicolumn{2}{|c|}{Leaky ReLU} \\
			\hline
			Convolution & $240\times320\times1$ & kernel=(3, 3) / strides=(1, 1) \\
			\hline
			Activation & \multicolumn{2}{|c|}{Sigmoidal} \\
			\hline
			Output & $240\times320\times1$ & - \\
			\hline	
		\end{tabular}
	}
\end{table}
The CNN proposed here introduces important modifications with respect to~\cite{dpdnet}. The most important change is
the use of spatially separable convolutions instead of separable
convolutions (depthwise convolution + pointWise convolution)~\cite{Xception}. The reason of this change lies in the fact that the separable convolutions isolate the depth channels in the phase of the depthwise convolution, and use
the pointwise convolution to synthetically increase the depth of the output. 

The operations performed by the separable convolutions can be harmful to the neural networks in some cases. The reason for this lies in the fact that especially the first layers of the convolutional neural network, where the characteristics of medium and low abstraction are found, in many cases complement each other. The break of these relationships or making them more complex by means of separable convolutional layers can
eliminate important relationships between the features that reduce the model training capacity or efficiency. In order to solve this problem without the need to resort to separable convolutional layers, we have used spatially separable convolutional layers, such as those used by Inception V3~\cite{inception}. An example of Spatially Separable
convolutions can be seen in Figure~\ref{fig:spatial_conv}.

Spatially separable convolutions use two 1D filters to compose a 2D equivalent convolution, so that it forces the filters to be decomposable in two 1D filters. This allows reducing the number of operations and parameters under certain conditions, which are shown in equations~\ref{eq:opt},~\ref{eq:opt1} and~\ref{eq:opt2}, where the variables $K$, $d$, $D$, $H$ and $W$ represent the kernel size, input depth of the tensor, final output depth of the tensor, height of the input tensor and width of the input tensor, respectively (as can be seen in Figure~\ref{fig:conv_spatial}).  

The following equations shows the conditions necessary to accomplish to reduce the number of operations and parameters, additionally, these equations suppose a stride of $1$ and a constant size of the image compensated with the padding.

\begin{equation}
\begin{gathered}
nparam_{conv}=KKdD=K^2dD \\
nops_{conv}=HWD(K^2d+(K^2-1)d)=HWD((2K^2-1)d)
\label{eq:opt}
\end{gathered}
\end{equation}

\begin{equation}
\begin{gathered}
nparam_{sep}=KdD+KD^2 \\
nops_{sep}=HWD(Kd+(K-1)d)+HWD(KD+(K-1)D)= \\
HWD((2K-1)(d+D)
\label{eq:opt1}
\end{gathered}
\end{equation}

\begin{equation}
\begin{gathered}
nparam_{conv}>nparam_{sep}\rightarrow K^2dD>KdD+KD^2\rightarrow\\
d>\frac{1}{(K-1)}D\\
nops_{conv}>nops_{sep}\rightarrow HWD((2K^2-1)d)>HWD((2K-1)(d+D))\\
d>\frac{(2K-1)}{2K(K-1)}D
\label{eq:opt2}
\end{gathered}
\end{equation}

Once we have defined the conditions, we apply them to the three
specific convolution cases used in the proposed CNN, where
$K=3,5,7$. The obtained values are shown in
table~\ref{tab:sep_conv_conditions} where its last row shows the more restrictive condition that allows improving both the number of parameters and the number of operations.

\begin{table}[htpb]
	\centering
	\begin{tabular}{|l c|c|c|c|}
		\hline 
		\multicolumn{2}{|c|}{Improvement in} & $K = 3$ & $K = 5$ & $K = 7$ \\   \hline \hline
		Parameters & $d>\frac{1}{(K-1)}D$        & $ d > 0.5D   $ & $ d > 0.25D  $ & $ d > 0.16D  $ \\  \hline
		Operations & $d>\frac{(2K-1)}{2K(K-1)}D$ & $ d > 0.416D $ & $ d > 0.225D $ & $ d > 0.154D $ \\
		\hline \hline
		\multicolumn{2}{|c|}{\textbf{Both}} & $ d > 0.5D $ & $ d > 0.25D $ & $ d > 0.16D $ \\ 
		\hline 
	\end{tabular}
	\caption{Conditions that the use of factorized
		convolutions must meet to be faster and more parameter efficient than
		conventional convolutions for the proposed CNN.} 
	\label{tab:sep_conv_conditions}
\end{table}

The main changes to the proposed architecture can be summarized as follows:
\begin{itemize}
	\item \textbf{Leaky Relu activation:}We use \emph{Leaky Relu} activations to improve the generalization and convergence, as we proposed in~\cite{activations}. \emph{Leaky Relu} helps to solve the "dying relu" problem, where the zero activation zone of the Relu slows and destabilizes the training process. Instead of the zero activation zone, the \emph{Leaky Relu} has a small negative slope that mitigates the problem.
	
	\item \textbf{Spatially separable convolutions:}In our proposal we use residual blocks with a certain basis of the ones used in~\cite{resnet16}, in terms of structure. The main difference is the inclusion of the spatially separable convolutions to speed up the network and solve the conventional separable convolution problems.
	
	\item \textbf{Resized convolutions:}In our proposal, we use the resized convolution as an approximation of the deconvolution process, with a nearest neighbor interpolation type. This is done to avoid the possible checkerboard artifacts that could appear in our output likelihood maps if we use the transposed convolution approximation, as discussed in~\cite{artifacts}. Additionally, we prefer the nearest neighbor interpolation instead of bilinear or bicubic interpolation, because they led to problems in the interpolation of high-frequency image features, as explained in~\cite{artifacts}.
	
	\item \textbf{Loss function:} An important change with respect to~\cite{dpdnet} is that the loss function used in the proposed system is adapted to the problem of detecting people with the Gaussian modeling approach used, so that it seeks to give greater weight to the learning of those Gaussians against the learning of the background of the image, which favors the convergence of the system, which was strongly affected by this factor.
	
\end{itemize}

\subsubsection{Architecture of the Main Block (\MB)}
\label{sec:arqu-main-block}

The architecture of the \emph{Main Block} (\MB) is shown in
Table~\ref{tab:mainBlock}, describing all its layers with its
correspondent dimensions and parameters, where \textit{a}, \textit{b} and \textit{c} are the numbers of filters of the internal layers in the \textit{Encoding Convolutional Blocks} (\ECB{}s), \textit{Decoding Convolutional Blocks} (\DCB{}s), and \textit{Identity Blocks} (\IB{}s).

The first layer of the \MB{} uses the $240\times 320$ input depth image $\mathcal{I}$, and, through the residual blocks with spatially separable convolutions, it codifies and processes the input to finally deliver the first likelihood map $\mathcal{C}$. The encoder consists of \IB{}s and \ECB{}s, where the first ones process the input tensors with the same output size as the input block but with higher depth, while the second ones process the input and generates an output with half the input size and higher depth. As opposed to the encoder, the decoder consists of \IB{}s and \DCB{}s that increase the data size and decreases depth, to obtain an output with the same size as the input depth image $\mathcal{I}$.

Following the efficiency conditions previously defined for the spatially separable convolutions, these convolutions will only be included in the network elements suitable to be improved in complexity and speed.

Regarding the Main Block layers, we initially have an ordinary
convolutional layer. This first layer is not spatially separable because it would not benefit from reduced complexity and increased speed, as introducing factorized convolution here would imply a computation time $10.24$ times slower than a conventional convolution. After the first convolution, a Batch Normalization and a \emph{Leaky Relu} activation will be applied, to finally use a Max Pooling layer that will
obtain the maximum energy of the filters.

After these initial filtering layers, we start to use the residual blocks, three \ECB{}s and \IB{}s will be used sequentially for the encoding process. They will be followed by another three \DCB{}s and \IB{}s that will be used sequentially for the decoding process. The final layers of the decoder use \textit{Cropping}, \textit{ZeroPadding} and \textit{UpSampling} to accommodate the image size, to finally apply the last convolutional layer followed by a batch normalization and
sigmoidal activation.

\subsubsection{Architecture of the Hypothesis Reinforcement Block (\HRB)}
\label{sec:arqu-hr-block}

The Hypothesis Reinforcement Block (\HRB{}) preserve the basic structure
of~\cite{dpdnet}, and it is a smaller version of the \MB{} that uses
residual blocks and spatially separable convolutions too. The \HRB{}
uses $\mathcal{I}_2$ as input, which is composed by $\mathcal{I}$ and
$\mathcal{C}$ with a size of $240\times 320\times 2$ as explained above.
In contrast to the proposal in~\cite{dpdnet} we have included an adder
at the output that adds the output \MB{} $\mathcal{C}$ to the \HRB{}
output and ensures that $\mathcal{C}$ is used as an initialization to
the polishing process, as the \HRB{} is in charge of obtaining the final
polished likelihood map $\mathcal{C}_{polished}$.  The \HRB{} is
composed firstly by the same structure as the \MB{}, it uses a first
convolutional layer, batch normalization, \emph{Leaky Relu} activation,
and max pooling.  After that, the number of \IB{}s, \ECB{}s, and \DCB{}s
is reduced to two for each type. Finally the output layers follow a
similar setup than the ones in the \MB{}, with a final sigmoidal
activation.  Table~\ref{tab:refBlock} summarizes all the layers and
blocks used in the \HRB{}. 

\begin{figure*}[!htbp]
	\centering
	\includegraphics[width=\linewidth]{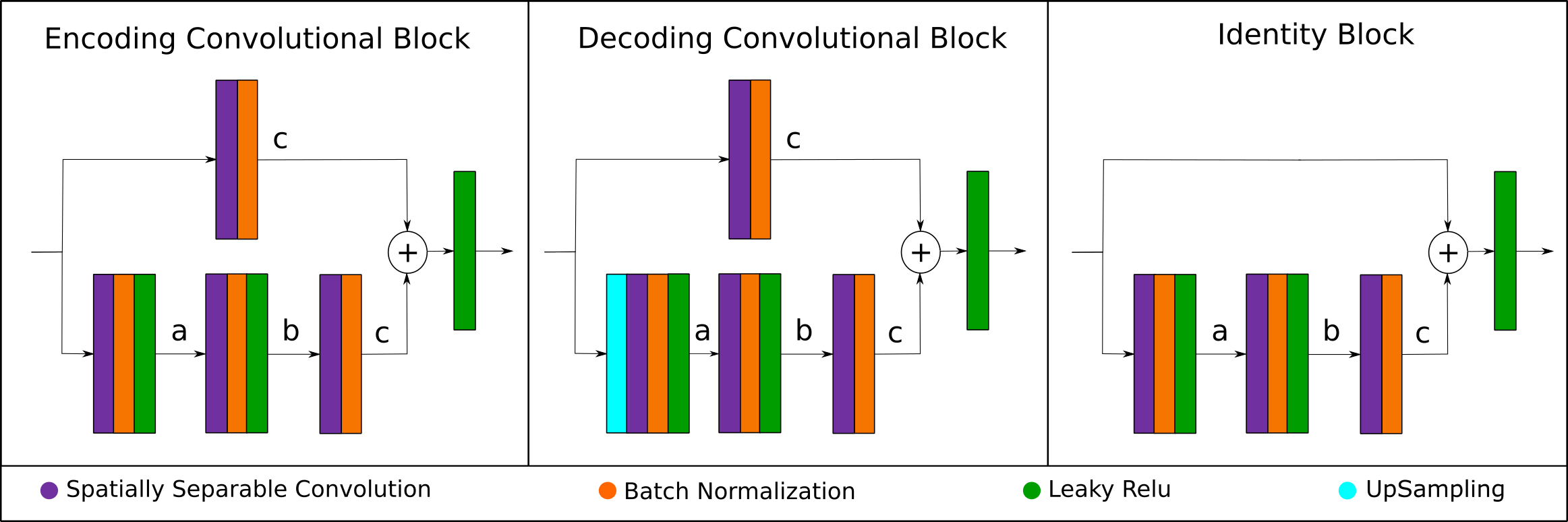}
	\caption{Architecture of the \ECB{}s, \DCB{}s and \IB{}s.}
	\label{fig:res-all}
\end{figure*}

\begin{table}[!htbp]
	\centering
	\caption{Detailed architecture of the hypothesis reinforcement block.}
	\label{tab:refBlock}
	\resizebox{0.5\columnwidth}{!}{
		\begin{tabular}{| l | c | c | }
			\hline
			\multicolumn{3}{|c|}{\textbf{Hypothesis Reinforcement Block (\HRB{})}} \\
			\hline
			\hline
			\multicolumn{1}{|c|}{\textbf{Layer}} & \textbf{Output size} & \textbf{Parameters} \\
			\hline
			\hline
			Input & $240\times320\times2$ & - \\
			\hline
			Convolution & $120\times160\times64$ & kernel=(7, 7) / strides=(2, 2) \\
			\hline
			BN & \multicolumn{2}{|c|}{-} \\
			\hline
			Activation & \multicolumn{2}{|c|}{Leaky ReLU} \\
			\hline
			Max Pooling & $40\times53\times64$ & size=(3, 3) \\
			\hline
			\hline
			\ECB{} & $40\times53\times256$ & \begin{tabular}{@{}c@{}}kernel=(3, 3) / strides=(1, 1) \\ (a=64, b=64, c=256) \end{tabular} \\
			\hline
			\ECB{} & $20\times27\times512$ & \begin{tabular}{@{}c@{}}kernel=(3, 3) / strides=(2, 2) \\ (a=128, b=128, c=512)  \end{tabular} \\
			\hline
			\DCB{} & $40\times54\times128$ & \begin{tabular}{@{}c@{}}kernel=(3, 3) / strides=(2, 2) \\ (a=512, b=512, c=128)  \end{tabular} \\
			\hline
			\DCB{} & $80\times108\times64$ & \begin{tabular}{@{}c@{}}kernel=(3, 3) / strides=(2, 2) \\ (a=256, b=256, c=64)  \end{tabular} \\
			\hline
			\hline
			Up Sampling & $240\times324\times64$ & size=(3, 3) \\
			\hline
			Cropping & $240\times320\times64$ & cropping=[(0, 0) (2, 2)] \\
			\hline
			Convolution & $240\times320\times64$ & kernel=(3, 3) / strides=(1, 1) \\
			\hline
			BN & \multicolumn{2}{|c|}{-} \\
			\hline
			Activation & \multicolumn{2}{|c|}{Leaky ReLU} \\
			\hline
			Convolution & $240\times320\times1$ & kernel=(3, 3) / strides=(1, 1) \\
			\hline
			Activation & \multicolumn{2}{|c|}{Sigmoidal} \\
			\hline
			Output  & $240\times320\times1$ & - \\
			\hline	
		\end{tabular}
	}
\end{table}
  The \ECB{}s and
\DCB{}s blocks are similar in structure. They are formed by two
unbalanced links, where the first one has three convolutional layers,
and the second has only one.  The output of the blocks is composed of
the added normalized output of the links.  The main difference between
the \ECB{}s and \DCB{}s lies in the convolutional process: the \ECB{}
uses a factorized convolution and the \DCB{} uses a resized convolution.
In the case of the \IB{}s, they preserve the size of the input, so that
one of the links only acts as a shortcut for the input to be directly
added to the output of the three convolutional links.  This preserves
the input information and adds them to the output filtered information
without changing the size.  The architecture of the \ECB{}s, \IB{}s and
\DCB{}s is shown in Figure~\ref{fig:res-all}, where the parameters
\textit{a}, \textit{b} and \textit{c} are the layer depth or the number
of filters in the corresponding layer.  The number of filters of the
third convolution in the bottom section must be equal to the number of
filters in the top section (\textit{c} parameter).  The parameters
\textit{a}, \textit{b} y \textit{c} in Tables~\ref{tab:mainBlock}
and~\ref{tab:refBlock} and in Figure~\ref{fig:res-all} have the same
meaning.

\subsection{Training procedure}
\label{sec:Training}

The network training process consists of two stages. In stage \emph{1}
we use a photorealistic synthetic database (created using the graphics
simulator~\cite{blender}) to train our CNN end to end. This allows the
network to learn the most important features for the person detection
task, while using a dataset that is automatically generated and
labeled. In stage \emph{2} we fine-tune the CNN to adjust its weights
to the actual camera pose with a small real database, composed of only
$3500$ frames that need to be manually labeled.  

The use of the synthetic database allows us to avoid the need to use or
create and label a large people detection database with depth images.
The synthetic database has $22\cdot10^3$ frames of depth images that
simulate capturing from an elevated front position around a synthetic
room\footnote{These images constitute the \DBGESDPD{} dataset, which has
	been made available to the scientific community~\cite{gesdpd2019}.}.

\begin{figure*}[!htbp]
	\centering
	\includegraphics[width=\textwidth]{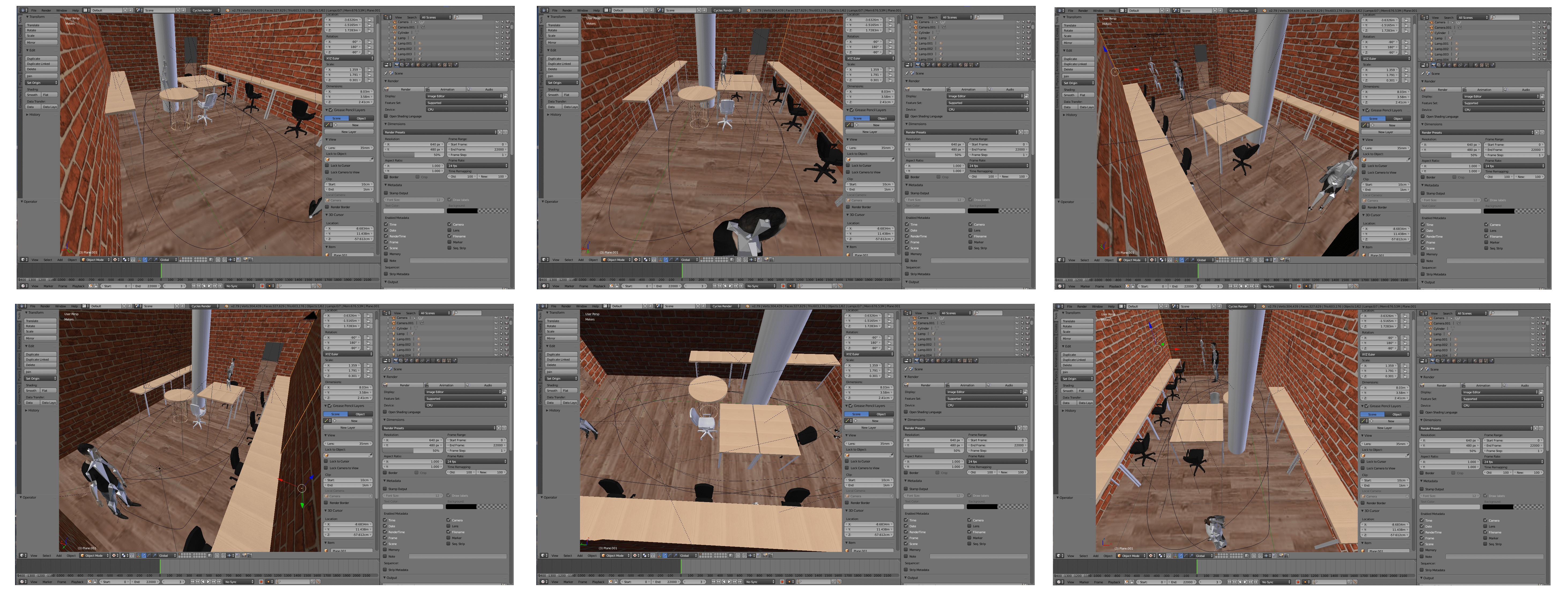}
	\includegraphics[trim={100 10 0 -10}, width=\textwidth]{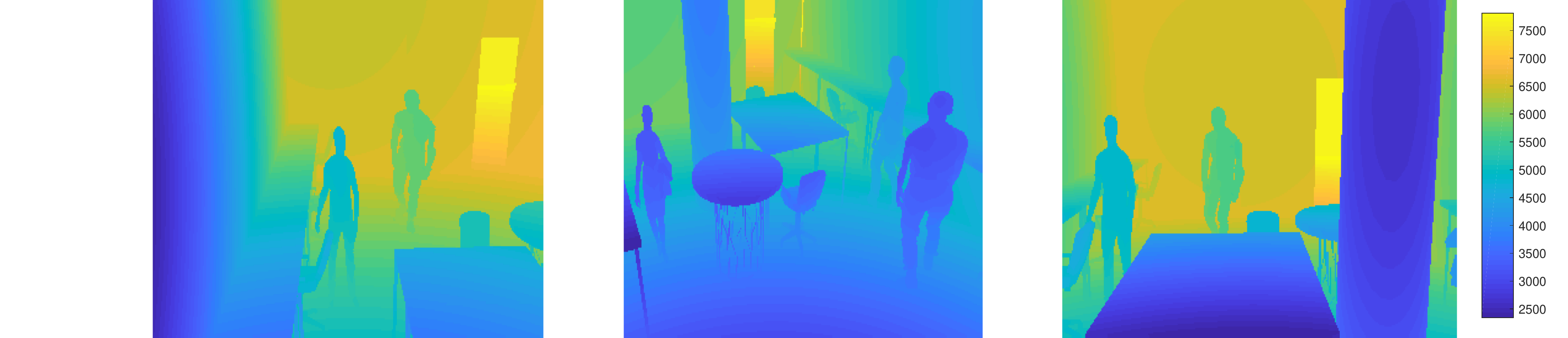}
	\caption{Sample images of the Blender simulated room with different
		perspectives (top two rows), and samples of synthetic depth images
		for training (belonging to the
		\DBGESDPD{} dataset, botom row).}
	\label{fig:sala}
\end{figure*}

In stage \emph{1} we split the
synthetic database in training and validation subsets composed by the
67\% and the 33\% percent of the database, respectively.  The context
used to create the synthetic database is a room similar to a laboratory
with persons walking in different directions as we can see in the first
two rows of Figure~\ref{fig:sala} as generated by Blender.  In addition
to this, we also changed the camera pose every frame, by rotating and
moving it around the room. Our objective is changing the background in
all the frames to prevent the CNN from learning a constant
background. This approach allows the network to more easily
generalize to different environments, as the CNN interprets the
background as noise and focus in the people walking around, isolating
the system from a given fixed perspective and pose of the
camera.

The images generated by the simulation software have a resolution of
$240\times320$, to be consistent with the images recorded with the
camera in the real scenario.  An example of the synthetically generated
depth images can be seen in the bottom row of Figure~\ref{fig:sala}.


In stage \emph{2} we used a recorded and manually labeled real database
composed by $3500$ frames recorded using a realsense D435
camera~\cite{realsense}. It was used fine-tune and adapt the network
weights network to the real environment.  This stage is necessary because
the synthetic data do not consider problems that appears in a real
scenario, such as motion blur, measurement noise or the influence of the
ambient light in the depth measures.

Regarding the labeling process, and taking into account that the input
image $\mathcal{I}$, the first likelihood map $\mathcal{C}$ and the
polished likelihood map $\mathcal{C}_{polished}$ are normalized between
$[0,1]$, we placed 2D Gaussian-like distributions in the centroid of the
labeled people with a maximum value of 1 corresponding to the
centroid. The standard deviation of the distributions is constant in all
the cases because using variable gaussian deviation proved to decrease
the detection performance in preliminary experiments. The gaussian
deviation value has been calculated using the average estimate of a
human head diameter, which is $\mathcal{D}=15$ pixels. Gaussian
deviation is calculated as $\sigma = \mathcal{D}/2.5=15/2.5=6$.

\begin{figure}[ht]
	\centering
	\includegraphics[trim={0 5cm 0 2cm},width=\columnwidth]{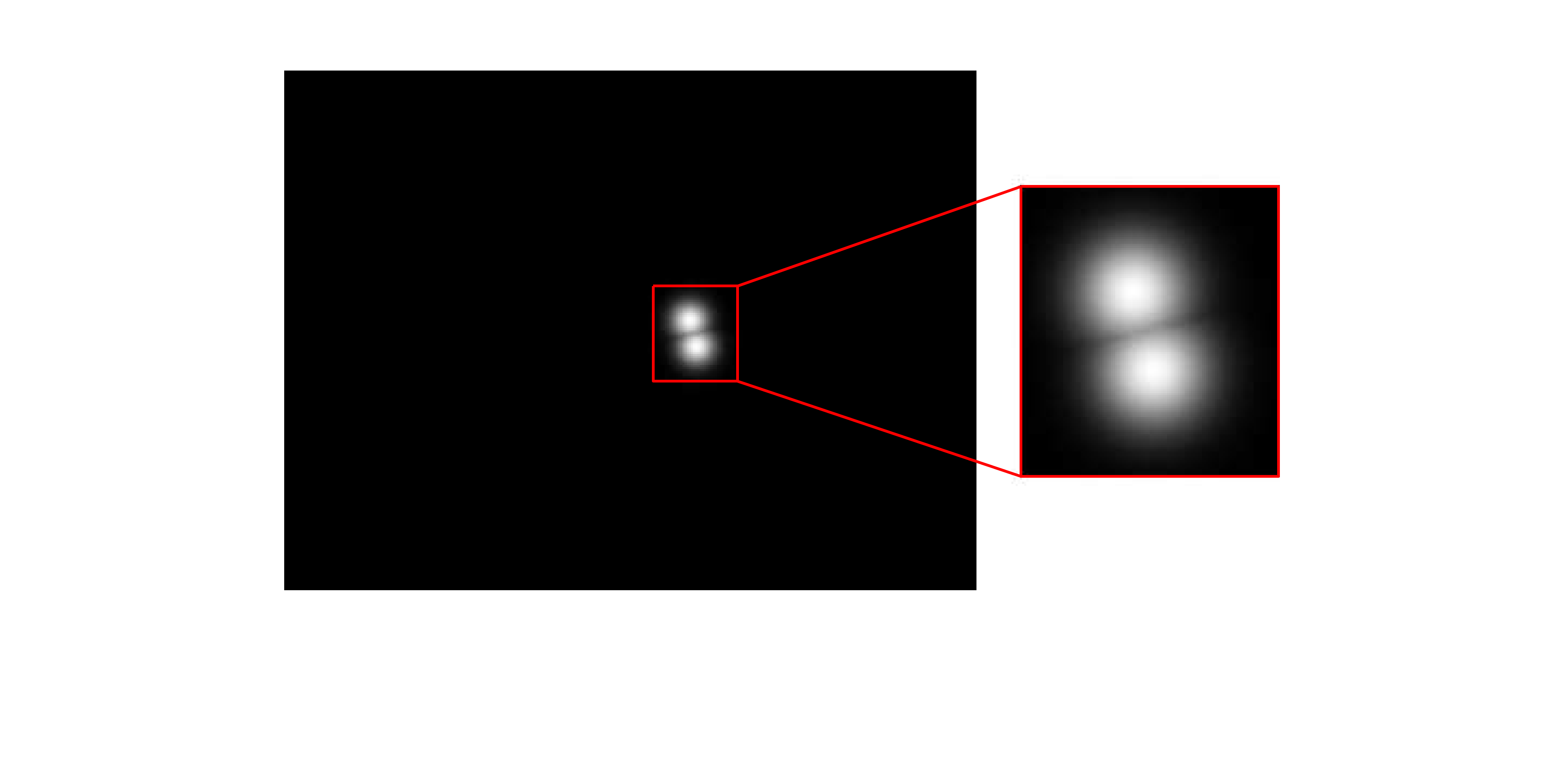}
	\caption{Gaussians Overlap Example.}
	\label{fig:gauss}
\end{figure}

One important modification on the labeling process as compared with the
proposal in~\cite{dpdnet} is that the gaussians are not constant in
terms of overlapping. Their standard deviations are constant, but when
two gaussians overlap, we force a clear separation between them, as
shown in Figure~\ref{fig:gauss}. This helps to mitigate the occlusions
problem because there always exists a separation between gaussians,
which in the post-processing will help to identify the people of the
image.  One example of how we handle the labeling in case of gaussian
overlap can be seen in Figure~\ref{fig:gauss}.

The optimization strategy we applied was \textit{Swats}, proposed
by~\cite{swats}. It involves a first training phase with
\textit{Adam}~\cite{kingma2014adam} corresponding to stage \emph{1},
and a second phase with \textit{SGD+Momemtum} that corresponds to stage
\emph{2}. This strategy allows to improve the network generalization
capabilities, as demonstrated in~\cite{swats}. The initial
\textit{learning rate} of the first stage was $0.001$, which was
combined with the use of an early stopping callback to save the best
possible model. The adaptive skills of \emph{Adam} modifies this
learning rate along the training process. In the second stage we use
the \textit{SGD+Momemtum} learning rate equal to $1e-5$.

The loss function has also been strongly improved as compared
with~\cite{dpdnet}, in which the simple mean square error was used. Our
CNN uses a training set composed of input depth images
$\mathcal{I}_i$ and initial and polished likelihood maps
$[\mathcal{C}_i,\mathcal{C}^{polished}_i]$. 

\begin{figure}[ht]
	\centering
	\includegraphics[width=0.8\columnwidth]{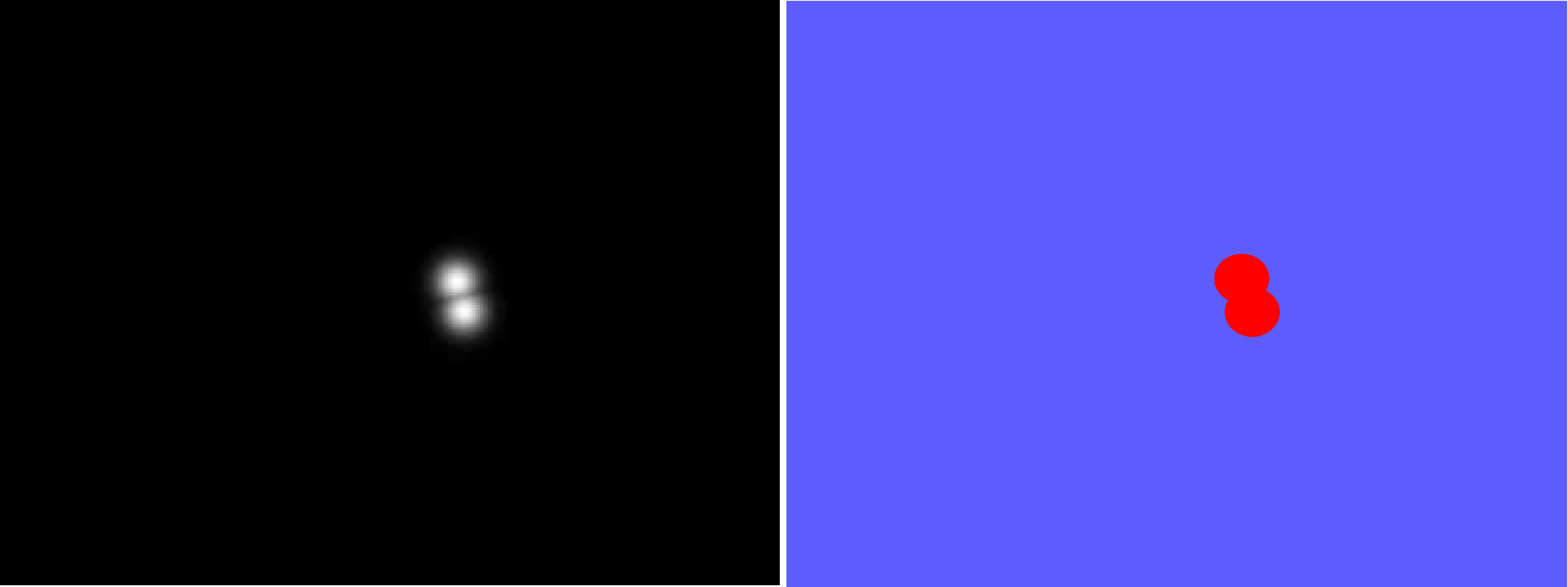}
	\caption{Example image of zero and non-zero points. In the left image, we can see an output likelihood map. Meanwhile in the right image we can observe the zero and positive-non-zero points plotted in blue and red respectively.}
	\label{fig:zerononzero}
\end{figure}

The proposed loss function tries to go beyond the conventional mean
square error function using a customized weighting in the detection and
background points, so that the system converges faster than in
conventional function cases. The proposed loss function consists of 4
elements, which are differentiated by location and type of points to be
evaluated. In the first case, we can separate in main's block output
location that its composed by the two first elements and the refinement
output location that it's composed of the third and fourth elements and
have a superindex named as $polished$. In the latter case, the loss
includes a weighting factor between the zero-negative value points
marked by a $0$ symbol and the positive-non zero points of the
likelihood map marked by a $+$ symbol. The first case helps to
refine the likelihood maps provided by the main block while the second
case improves the convergence to the gaussian distributions of the
target likelihood map. The loss function is described in
equation~\ref{eq:1}.

\begin{equation}
\begin{gathered}
\mathcal{L} = \lambda_1\frac{1}{N}\sum_{i=1}^N \|\hat{\mathcal{C}_{i_+}} - \mathcal{C}_{i_+}\|^2 + \lambda_2 \frac{1}{N}\sum_{i=1}^N \|\hat{\mathcal{C}_{i_0}} - \mathcal{C}_{i_0}\|^2+\\\lambda_1\frac{1}{N}\sum_{i=1}^N \|\hat{\mathcal{C}^{polished}_{i_+}} - \mathcal{C}^{polished}_{i_+}\|^2 + \lambda_2 \frac{1}{N}\sum_{i=1}^N \|\hat{\mathcal{C}^{polished}_{i_0}} - \mathcal{C}^{polished}_{i_0}\|^2,
\label{eq:1}
\end{gathered}
\end{equation}

\noindent where $\lambda_2=1$ and $\lambda_1=1.3$ are the parameters
that weight the importance of the zero-negative value points and the 
non-zero-positive points in the loss function. An example 
of the zero and non-zero points considered can be seen in 
Figure~\ref{fig:zerononzero}, where at the left side we have the original 
likelihood map and at the right in red we distinguish the non-zero points 
and in blue the zero points.

We trained the network for 30 epochs with the \emph{Adam} Optimizer in
the first stage, and for 20 epochs with \emph{SGD+Momemtum} in the
second stage, choosing the best possible model obtained along the
training process.

\section{Experimental Work}
\label{sec:experimental-work}

This section first describes the datasets (section~\ref{sec:datasets})
and the experimental setup (section~\ref{sec:experimental-setup}).
Then, we present the evaluated algorithms
(section~\ref{sec:evaluated-algorithms}), and the used metrics
(section~\ref{sec:evaluation-metrics}).
Finally, we provide details on the training and hyperparameter selection 
strategies (section~\ref{sec:thresh-select-strat}) and the way to generate 
part of the metrics for the algorithm proposed here 
(section \ref{sec:roc-curv-gener}). 

Once the experimental conditions and the data to be used are described,
the results are presented through the use of tables and figures, in
which the main metrics are detailed and a comparison between our
proposal and other algorithms in the literature is carried out. The
results are first presented for the synthetic data case (section
~\ref{sec:results-train-with-synth-data}) which allows us to introduce
and explain in-depth the effectiveness of a synthetic pre-training as
the one performed here, and which are the limitations that can be found
in the real world. After explaining the synthetic training, the real
results are presented in each of the databases used (section
~\ref{sec:results-dataset-specific-modelsdiscussion}).  This explanation
ends with a comparison of the average results of all the databases
(section \ref{sec:average-results-all-datasets-specific}) and a
discussion about the use of a global training approach to the system
(section \ref{sec:disc-prec-recall}).  After the presentation of the
results, the computational performance is then evaluated quantitatively
(section \ref{sec:perf-eval}).

\subsection{Datasets \& Data Partition}
\label{sec:datasets}

\begin{figure*}[!htbp]
	\begin{center}
		\begin{subfigure}[t]{0.88\textwidth}
			\includegraphics[width=0.49\textwidth]{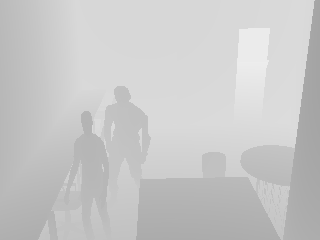}\;\includegraphics[width=0.49\textwidth]{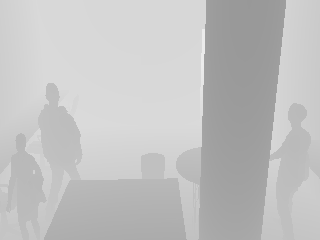}
			\caption{Sample frames from the GESDPD dataset~\cite{gesdpd2019}.}
			\label{fig:gesdpd-samples}
		\end{subfigure}
		\begin{subfigure}[t]{0.88\textwidth}
			\includegraphics[width=0.49\textwidth]{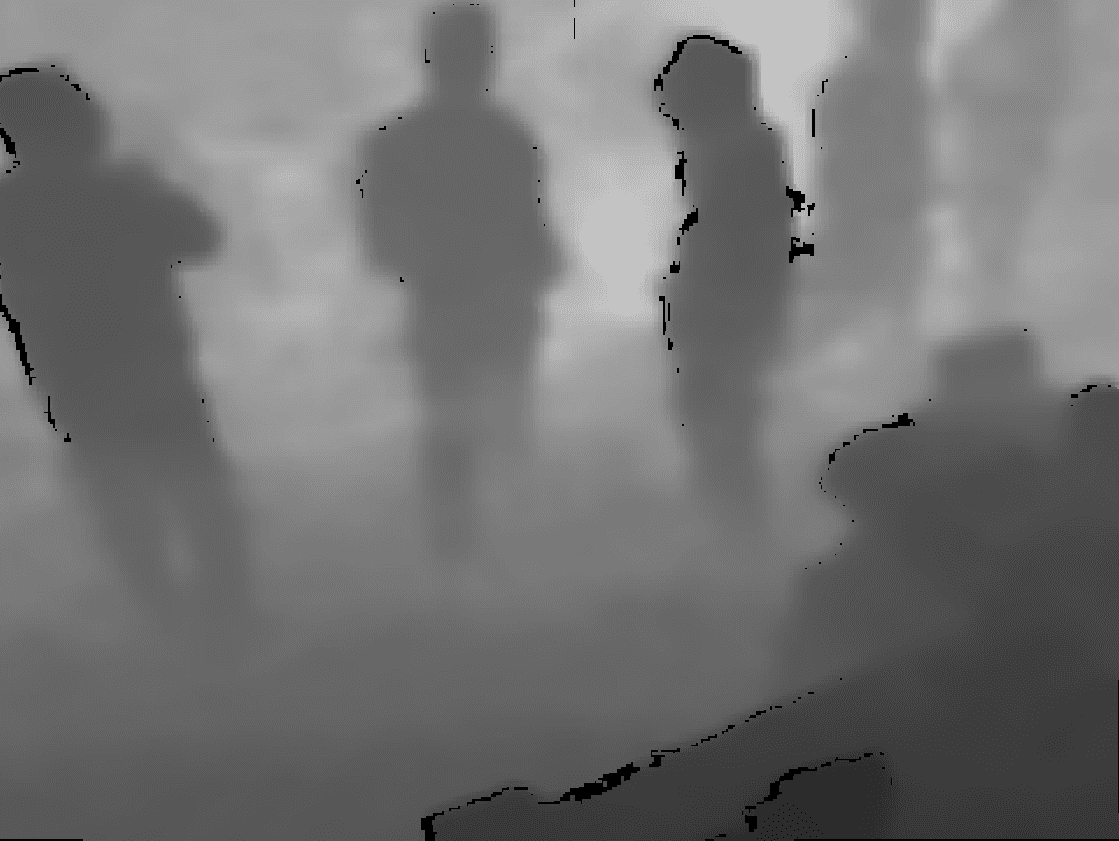}\;\includegraphics[width=0.49\textwidth]{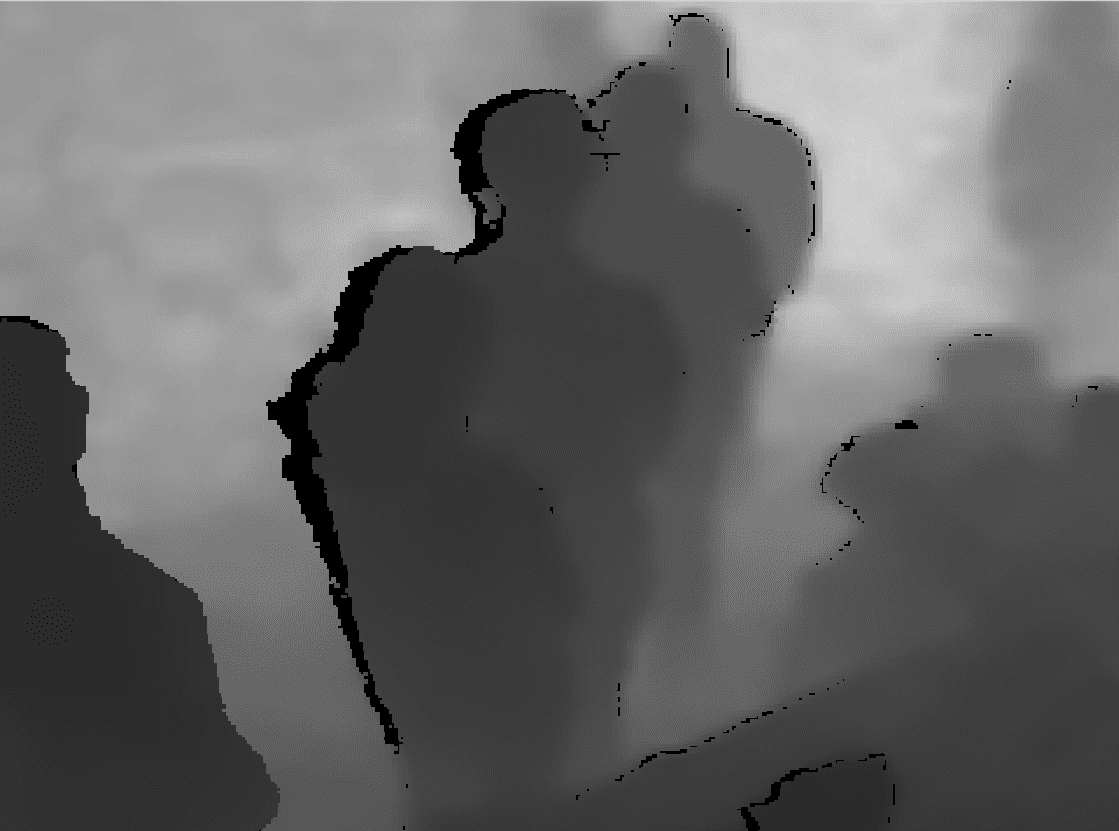}
			\caption{Sample frames from the GFPD dataset~\cite{GFPD}.}
			\label{fig:gfpd-samples}
		\end{subfigure}
		\begin{subfigure}[t]{0.88\textwidth}
			\includegraphics[width=0.49\textwidth]{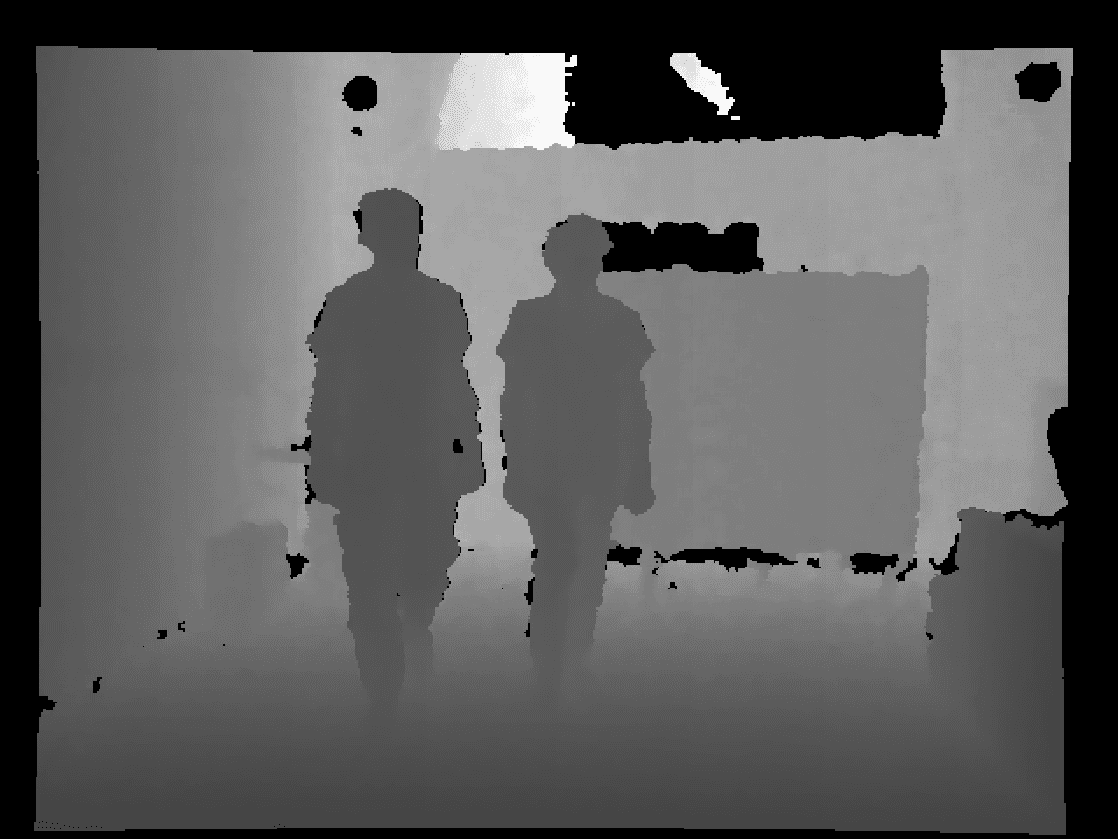}\;\includegraphics[width=0.49\textwidth]{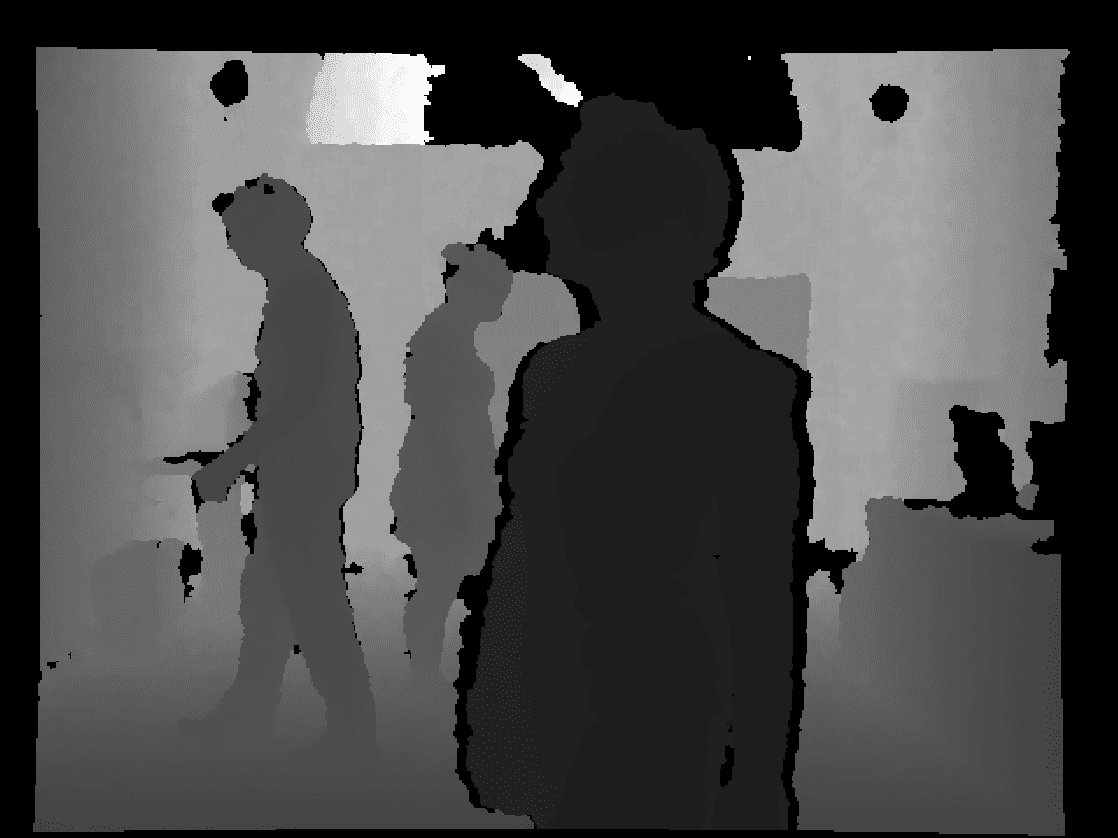}
			\caption{Sample frames from the KTP dataset~\cite{munaro2014}.}
			\label{fig:ktp-samples}
		\end{subfigure}
	\end{center}
	\caption{Sample frames from the GESDPD, GFPD and KTP databases (gray coded depth).}
	\label{fig:sample-frames1}
\end{figure*}

\begin{figure*}[!htbp]
	\begin{center}
		\begin{subfigure}[t]{0.85\textwidth}
			\includegraphics[width=0.49\textwidth]{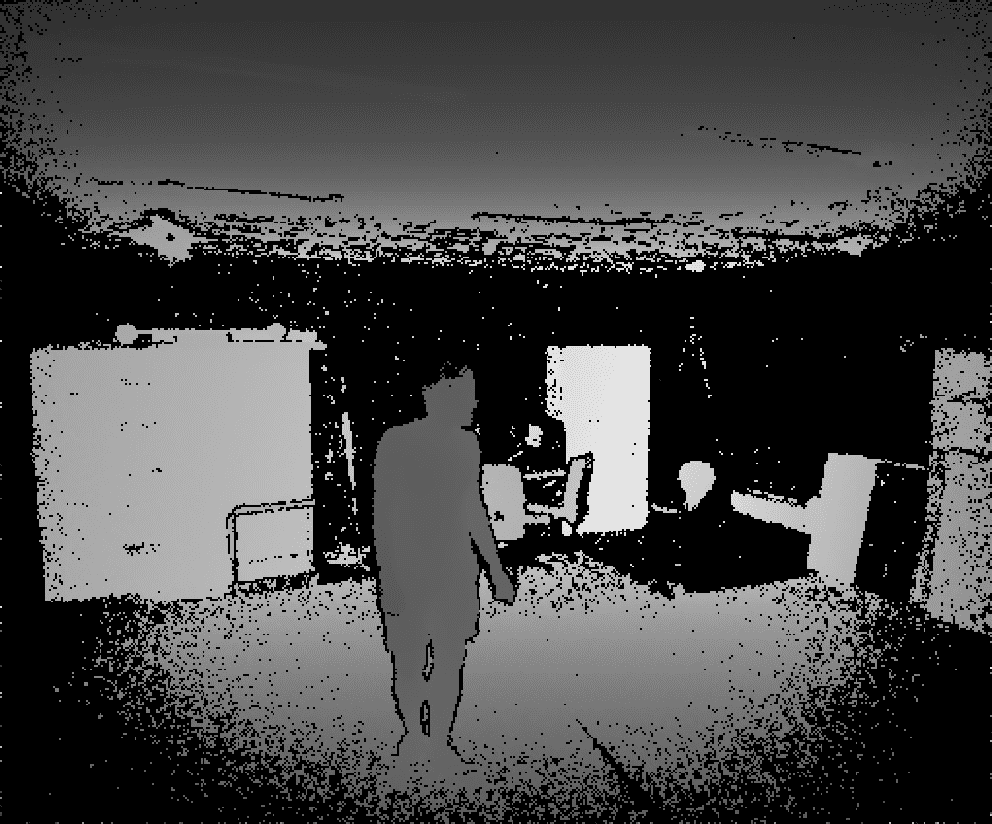}\;\includegraphics[width=0.49\textwidth]{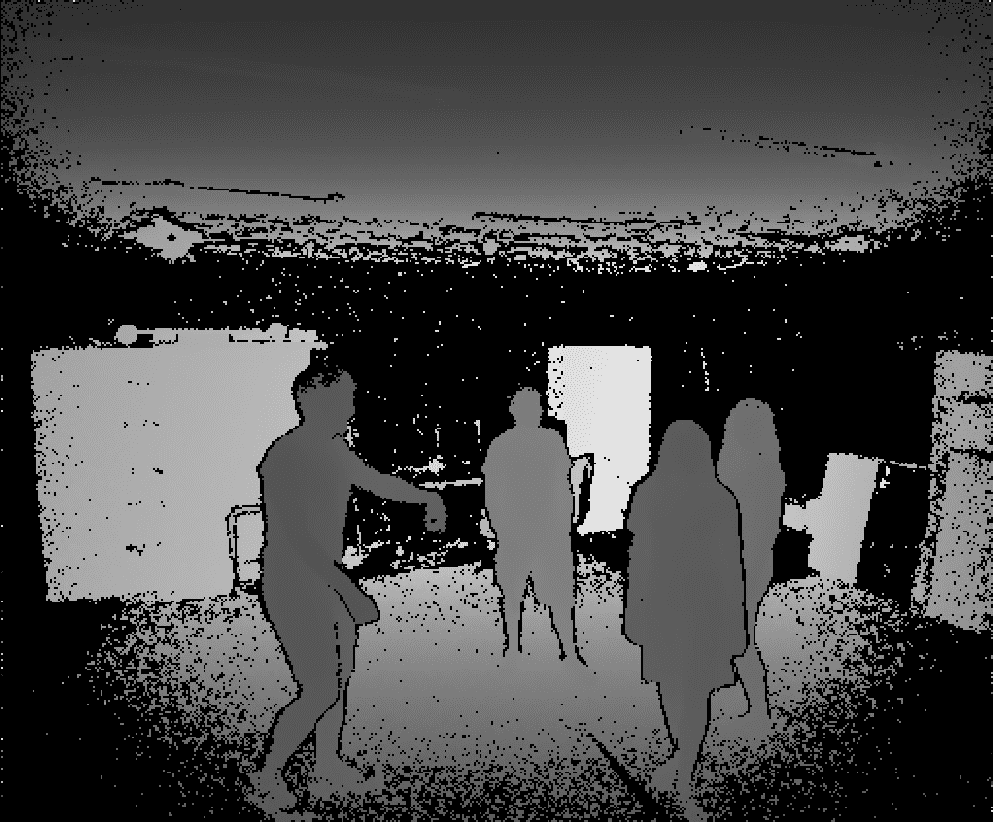}
			\caption{Sample frames from the EPFL-LAB dataset~\cite{Bagautdinov_CVPR_2015_dpom}.}
			\label{fig:epfl-lab-samples}
		\end{subfigure}
		\begin{subfigure}[t]{0.85\textwidth}
			\includegraphics[width=0.49\textwidth]{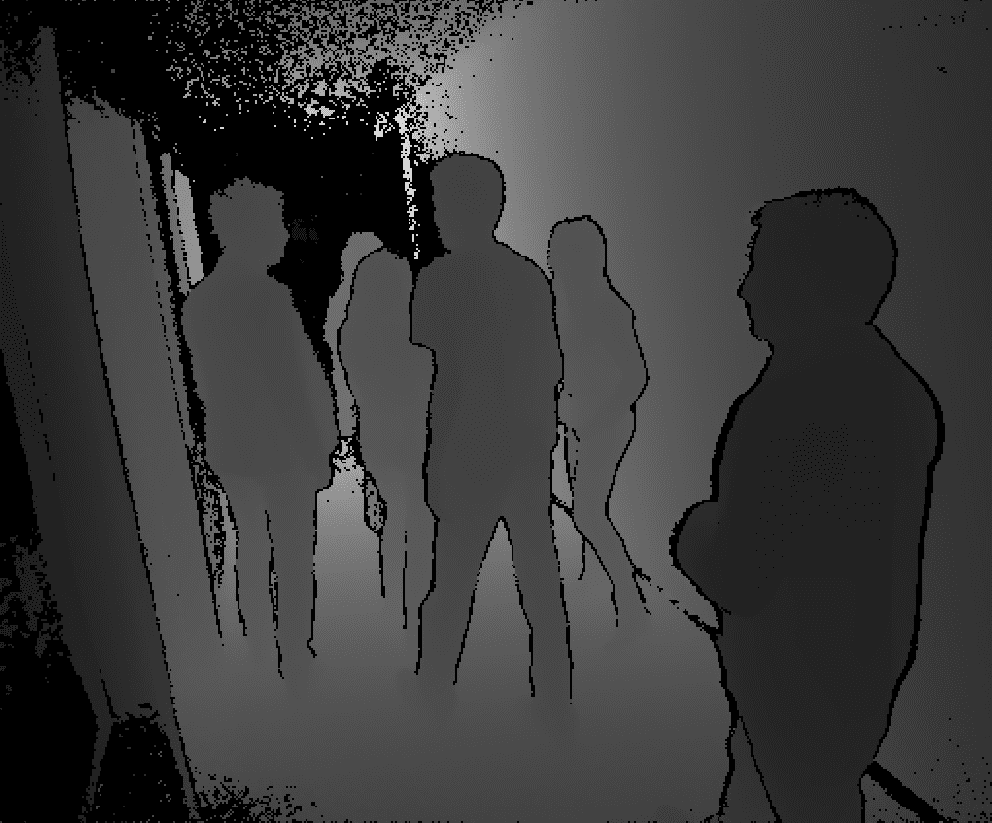}\;\includegraphics[width=0.49\textwidth]{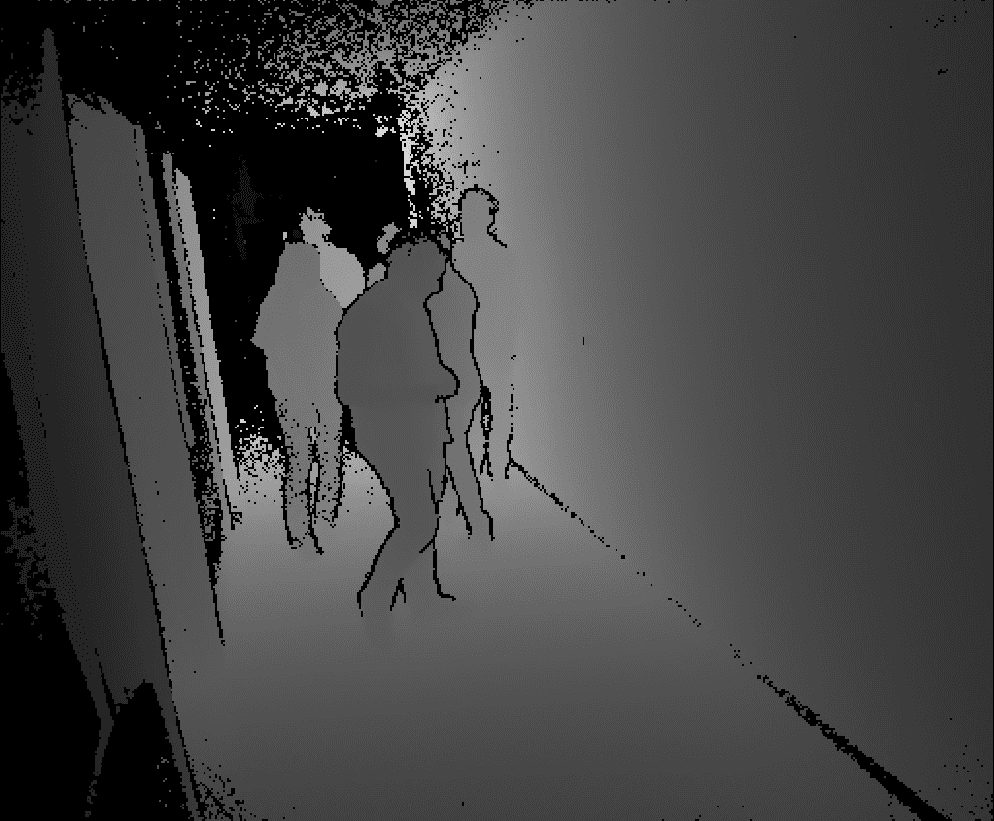}
			\caption{Sample frames from the EPFL-CORRIDOR dataset~\cite{Bagautdinov_CVPR_2015_dpom}.}
			\label{fig:epfl-corridor-samples}
		\end{subfigure}
		\begin{subfigure}[t]{0.85\textwidth}
			\includegraphics[width=0.49\textwidth]{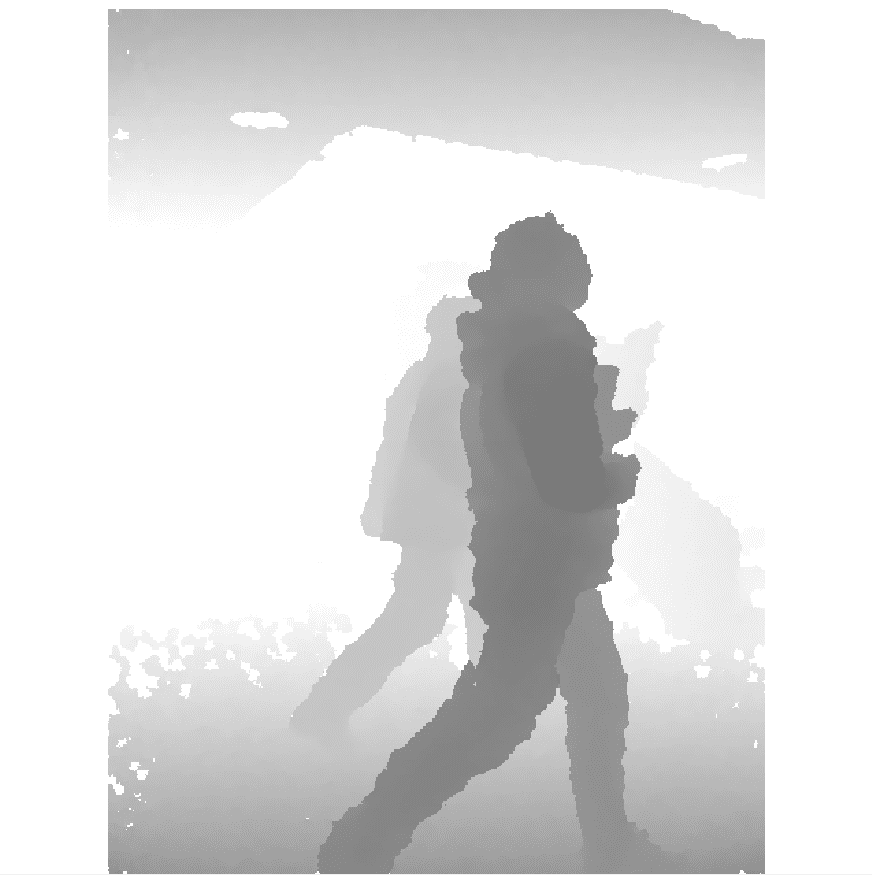}\;\includegraphics[width=0.49\textwidth]{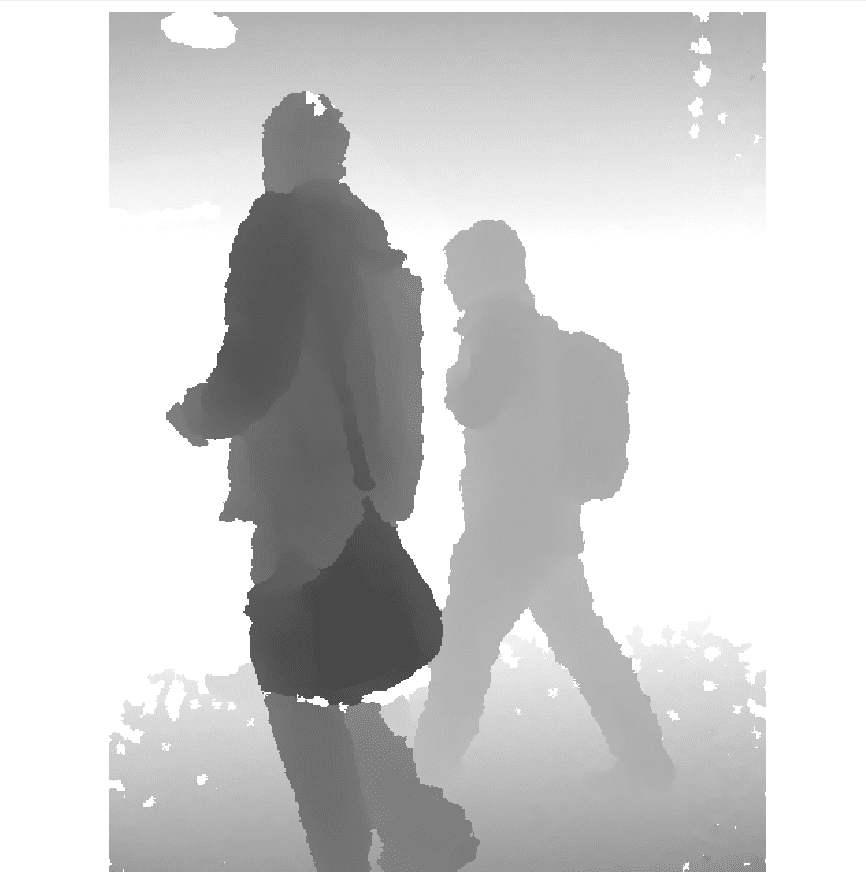}
			\caption{Sample frames from the UNIHALL dataset~\cite{unihall}.}
			\label{fig:unihall-samples}
		\end{subfigure}
	\end{center}
	\caption{Sample frames from the EPFL-LAB, EPFL-CORRIDOR and UNIHALL databases (gray coded depth).}
	\label{fig:sample-frames2}
\end{figure*}

In order to provide a wide range of evaluation conditions, we have used
five different databases, that will be described
next. Figures~\ref{fig:sample-frames1} and~\ref{fig:sample-frames2}
provide some sample frames to give an idea on the style and quality of
the different datasets. 

\begin{enumerate}
	\item \textbf{GESDPD}~\cite{gesdpd2019}: GESDPD is a synthetic dataset
	which contains 22000 depth images, that simulate to have been taken
	with a sensor in an elevated front position, in an indoor working
	environment. These images have been generated using the simulation software
	Blender~\cite{blender}. The simulated scene shows a room with
	different people walking in different directions.  The camera
	perspective is not stationary, as it rotates and moves along the
	dataset, which avoids a constant background that could be learned by
	CNN in the training, as can be seen in figure~\ref{fig:sala}, that
	shows different perspectives of the synthetic room in the simulation
	software Blender~\cite{blender}.  Using different backgrounds around
	the synthetic room allows CNN to see the background as noise and focus
	the training in the people that come along the image, immunizing the
	network to the change of camera perspective and assembly conditions.
	
	The generated images have a resolution of $320\times240$ pixels
	codified as 16 bits unsigned integers. Some examples of the synthetic
	images are shown in figure~\ref{fig:sala}, the images correspond to
	three different perspectives, and the depth values are represented
	using a colormap.

	\item \textbf{GFPD}~\cite{GFPD}: The Geintra Frontal Person Detection dataset is a
	high-resolution dataset recorded with a \cite{realsense} realsense D435
	camera. GFPD contains 5270 depth frames with 13827
	annotated people instances. The camera has an active stereo depth
	sensor that provides depth maps with a resolution of
	$1280\times720$. The records of this dataset consider a great variety
	of conditions including different sensor heights (2200-2700 mm),
	different tilt angles (26-41 degrees), as well as different
	backgrounds and lighting conditions.
	
	\item \textbf{EPFL}~\cite{Bagautdinov_CVPR_2015_dpom}, that includes two different datasets:
	\begin{enumerate}
		\item \textbf{EPFL-LAB:} The first one (EPFL-LAB) contains around
		1000 RGB-D frames with around 3000 annotated people
		instances. There are at most 4 people who are mostly facing the
		camera, presumably a scenario for which the Kinect software was
		fine-tuned. 
		
		\item \textbf{EPF-CORRIDOR:} The second one (EPFL-CORRIDOR) was
		recorded in a more realistic environment, a corridor in a
		university building. It contains over 3000 frames with up to 8
		individuals, and it is a challenging dataset since there are
		important occlusions.
	\end{enumerate}
	
	\item \textbf{KTP}~\cite{munaro2014}: The Kinect Tracking Precision
	dataset (KTP) presented in~\cite{munaro2014} contains several
	sequences of at most 5 people walking in a small lab environment. They
	were recorded by a depth camera mounted on a robot platform and we use
	here the only one that was filmed while the camera was static. Authors
	provide ground truth locations of the individuals both on the image
	plane, and on the ground plane. Unfortunately, the quality of the
	ground truth for the ground plane is limited, due to the poor quality
	registration of the depth sensor location to the environment. In order
	to fix this, we made an effort and manually specified points
	corresponding to individuals on the depth maps, then projected them on
	the ground plane, and took an average to get a single point
	representing person location. This introduces a small bias as we only
	observe the outer surface of the person but any motion capture system
	would have similar issues.
	
	\item \textbf{UNIHALL}~\cite{unihall}: In \cite{unihall, unihall2},
	the authors report their results in a dataset containing about 4500
	RGB-D images recorded in a university hall from three statically
	mounted Kinect cameras. Unfortunately, there is no ground plane ground
	truth available, thus we only report results for image plane. To
	compare to their results, we follow the evaluation procedure described
	in~\cite{unihall}, that is, without penalizing approaches for not
	detecting occluded or hidden people. We also report our performance
	for the full dataset separately.
	
\end{enumerate}

In what respect to the data partitioning we did to generate the training
and testing subsets, Table~\ref{tab:data-partitioning} shows the total
number of frames (column \emph{\#framesFull}), the total number of
people labeled in the ground truth in these frames (column
\emph{\#PeopleFull}), and the partition statistics for the training
subsets (columns \emph{\#framesTrain} and \emph{\#PeopleTrain}), and for
the testing subsets (columns \emph{\#PeopleTest} and
\emph{\#framesTest}). It also provides details, wherever necessary, on
the sequences used. Total numbers are shown right-aligned, and when the
partition involved several sequences, the corresponding data for
individual sequences is shown left-aligned. To give an idea on the
relative size of the training and testing material for each dataset,
Table~\ref{tab:data-partitioning} also provides the percentages
corresponding to these subsets in the rows with accumulated totals for
the \emph{\#PeopleTrain} and \emph{\#PeopleTest} columns. 

\begin{table*}[ht]
	\caption{Data partition details.}
	\label{tab:data-partitioning}
	\resizebox{\textwidth}{!}{%
		\begin{tabular}{|c|l|l|l|l|l|l|l|}
			\hline
			& Sequence               & \#framesFull & \#PeopleFull & \#FramesTrain            & \#PeopleTrain & \#FramesTest             & \#PeopleTest \\ \hline\hline
			\DBGESDPD                         &                        & \multicolumn{1}{r|}{\textbf{22000}}        & \multicolumn{1}{r|}{\textbf{54215}}        & \multicolumn{1}{r|}{\textbf{17600}} & \multicolumn{1}{r|}{\textbf{36063}}    & \multicolumn{1}{r|}{\textbf{4400}} & \multicolumn{1}{r|}{\textbf{18152}}        \\ [-10pt]
			&                        &                                            &                                            &                                     & \multicolumn{1}{r|}{\textbf{(67\%)}}    &                                    & \multicolumn{1}{r|}{\textbf{(33\%)}}        \\ \hline\hline
			\multirow{6}{*}{\DBGFPD{}}        & 1                      & 1690         & 4462         & 1690                     & 4462          &                          &              \\ \cline{2-8} 
			& 2                      & 900          & 3760         & 900                      & 3760          &                          &              \\ \cline{2-8} 
			& 3                      & 1090         & 4337         &                          &               & 1090                     & 4337         \\ \cline{2-8} 
			& 4                      & 500          & 1268         & 500                      & 1268          &                          &              \\ \cline{2-8} 
			& \multicolumn{1}{r|}{\textbf{Total \DBGFPD}} & \multicolumn{1}{r|}{\textbf{4180}}         & \multicolumn{1}{r|}{\textbf{13827}}        & \multicolumn{1}{r|}{\textbf{3090}}                     & \multicolumn{1}{r|}{\textbf{9490}}          & \multicolumn{1}{r|}{\textbf{1090}}                     & \multicolumn{1}{r|}{\textbf{4337}}         \\ [-10pt]
			&                        &                                            &                                            &                                     & \multicolumn{1}{r|}{\textbf{(69\%)}}    &                                    & \multicolumn{1}{r|}{\textbf{(31\%)}}        \\ \hline\hline
			\DBEPFLLAB                        & 1                      & \multicolumn{1}{r|}{\textbf{920}}          & \multicolumn{1}{r|}{\textbf{2287}}         & \multicolumn{1}{r|}{\textbf{600}}                      & \multicolumn{1}{r|}{\textbf{1385}}          & \multicolumn{1}{r|}{\textbf{320}}                      & \multicolumn{1}{r|}{\textbf{902}}          \\ [-10pt]
			&                        &                                            &                                            &                                     & \multicolumn{1}{r|}{\textbf{(61\%)}}    &                                    & \multicolumn{1}{r|}{\textbf{(39\%)}}        \\ \hline\hline
			\multirow{7}{*}{\DBEPFLCORRIDOR}  & 20141008\_141829\_00   & 390          & 899          & 390                      & 899           &                          &              \\ \cline{2-8} 
			& 20141008\_1414\_30\_00 & 420          & 813          & 420                      & 813           &                          &              \\ \cline{2-8} 
			& 20141008\_141913\_00   & 396          & 1886         & 396                      & 1886          &                          &              \\ \cline{2-8} 
			& 20141008\_141537\_00   & 430          & 853          &                          &               & 430                      & 853          \\ \cline{2-8} 
			& 20141008\_141537\_00   & 1100         & 3581         &                          &               & 1100                     & 3581         \\ \cline{2-8} 
			& \multicolumn{1}{r|}{\textbf{Total \DBEPFLCORRIDOR}} & \multicolumn{1}{r|}{\textbf{2736}}         & \multicolumn{1}{r|}{\textbf{8032}}         & \multicolumn{1}{r|}{\textbf{1206}}                     & \multicolumn{1}{r|}{\textbf{3598}}          & \multicolumn{1}{r|}{\textbf{1530}}                     & \multicolumn{1}{r|}{\textbf{4434}}         \\ [-10pt]
			&                        &                                            &                                            &                                     & \multicolumn{1}{r|}{\textbf{(45\%)}}    &                                    & \multicolumn{1}{r|}{\textbf{(55\%)}}        \\ \hline\hline
			\multirow{3}{*}{\DBKTP}           & ROTATION               & 2200         & 3299         & 2200                     & 3299          &                          &              \\ \cline{2-8} 
			& STILL                  & 2100         & 3420         &                          &               & 2100                     & 3420         \\ \cline{2-8} 
			& \multicolumn{1}{r|}{\textbf{Total \DBKTP}} & \multicolumn{1}{r|}{\textbf{4300}}         & \multicolumn{1}{r|}{\textbf{6719}}         & \multicolumn{1}{r|}{\textbf{2200}}                     & \multicolumn{1}{r|}{\textbf{3299}}          & \multicolumn{1}{r|}{\textbf{2100}}                     & \multicolumn{1}{r|}{\textbf{3420}}         \\ [-10pt]
			&                        &                                            &                                            &                                     & \multicolumn{1}{r|}{\textbf{(49\%)}}    &                                    & \multicolumn{1}{r|}{\textbf{(51\%)}}        \\ \hline\hline
			\DBUNIHALL                        & mensa\_seq0\_1.1       & \multicolumn{1}{r|}{\textbf{2900}}         & \multicolumn{1}{r|}{\textbf{2979}}         & \multicolumn{1}{r|}{\textbf{1400}}                     & \multicolumn{1}{r|}{\textbf{2054}}          & \multicolumn{1}{r|}{\textbf{1500}}                     & \multicolumn{1}{r|}{\textbf{925}}          \\ [-10pt]
			&                        &                                            &                                            &                                     & \multicolumn{1}{r|}{\textbf{(69\%)}}    &                                    & \multicolumn{1}{r|}{\textbf{(31\%)}}        \\ \hline\hline
			\textbf{ALL Datasets}                       &  \multicolumn{1}{r|}{\textbf{Total ALL Datasets}}     & \multicolumn{1}{r|}{\textbf{37036}}         & \multicolumn{1}{r|}{\textbf{88059}}         & \multicolumn{1}{r|}{\textbf{26096}}                     & \multicolumn{1}{r|}{\textbf{55889}}          & \multicolumn{1}{r|}{\textbf{10940}}                     & \multicolumn{1}{r|}{\textbf{32170}}          \\ [-10pt]
			&                        &                                            &                                            &                                     & \multicolumn{1}{r|}{\textbf{(63\%)}}    &                                    & \multicolumn{1}{r|}{\textbf{(37\%)}}        \\ \hline
			
		\end{tabular}
	}
\end{table*}
\subsection{Experimental setup}
\label{sec:experimental-setup}

\subsubsection{Evaluated algorithms}
\label{sec:evaluated-algorithms}

We compared the performance of our proposal \AlgDPDNetvtwo{} with up to
ten different strategies described in the literature that use
depth cameras with an elevated
non-overhead position in a people detection task.

Table~\ref{tab:sotaMethods} shows the main characteristics of all
the evaluated methods, considering both classical and DNN
approaches. It is relevant to note that we are comparing our depth-only
proposal with others using different combinations of RGB and depth
information, which imposes strong differences in the quality of the
exploited data.
\begin{table}[!htbp]
	\centering
	\caption{Evaluated state-of-the-art methods for multiple people detection.}
	\resizebox{0.6\textwidth}{!}{%
		\begin{tabular}{|c|c|c|c|}
			\hline
			Solutions & Methods & \begin{tabular}[c]{@{}c@{}}Input\\ information\end{tabular} & References \\ \hline\hline
			\cellcolor[HTML]{FFFFFF} & ACF & {\color[HTML]{000000} RGB} & \cite{acf}  \\ \cline{2-4} 
			\cellcolor[HTML]{FFFFFF} & PCL-Munaro & {\color[HTML]{000000} RGB-D} & \cite{munaro2014} \\ \cline{2-4} 
			\cellcolor[HTML]{FFFFFF} & Kinect2 & {\color[HTML]{000000} RGB-D} & \cite{kinect2}  \\ \cline{2-4} 
			\multirow{-4}{*}{\cellcolor[HTML]{FFFFFF}Classic} & Unihall & {\color[HTML]{000000} RGB-D} & \cite{unihall} \\ \hline\hline
			\cellcolor[HTML]{FFFFFF} & DPOM & {\color[HTML]{000000} D} & \cite{Bagautdinov_CVPR_2015_dpom} \\ \cline{2-4} 
			\cellcolor[HTML]{FFFFFF} & RCNN & {\color[HTML]{000000} RGB} & \cite{rcnn} \\ \cline{2-4} 
			\cellcolor[HTML]{FFFFFF} & RGBCNN & {\color[HTML]{000000} RGB} & \cite{mva2017} \\ \cline{2-4} 
			\cellcolor[HTML]{FFFFFF} & RGBCECDCNN & RGB-D & \cite{mva2017} \\ \cline{2-4} 
			\multirow{-5}{*}{\cellcolor[HTML]{FFFFFF}DNN} & YOLO V3 &  {\color[HTML]{000000} RGB} & \multicolumn{1}{l|}{\cite{yolov3}} \\ \hline
		\end{tabular}%
	}
	\label{tab:sotaMethods}
\end{table}
Below, we explain in detail the evaluated state-of-the-art methods we
use in the comparison:

\begin{itemize}
	
	\item \AlgACFRef{}: That uses a RGB detector from~\cite{acf}, based
	on AdaBoost and aggregate channel features~\cite{acf2} to give a
	sense of what a state-of-the-art detector that does not use depth can
	do on these sequences.
	
	\item \AlgPCLMunaroRef{}: That uses a RGB-D detector
	from~\cite{munaro2014}, based on modified HOG features on regions
	extracted by depth segmentation.
	
	\item \AlgKINECTtwoRef{}: Based on the results obtained from the human
	pose estimation of the latest Kinect for Windows
	SDK~\cite{kinect2}. It is not publicly known what specific algorithm
	is used. However in~\cite{kinect21}, the authors report that their
	algorithm is at the core of the human pose estimation for the older
	version of the Kinect software. For undisclosed reasons, the framework
	supports tracking up to 6 people, with the working depth range limited
	to 4.5 meters. To ensure fairness, we kept these restrictions in mind
	when using the \DBEPFLLAB{} and \DBEPFLCORRIDOR{} datasets. We do not
	penalize algorithms for not detecting more than 6 people or people who
	are further than 4.5 meters away.
	
	\item \AlgUNIHALLRef{}: That uses a RGB-D detector from~\cite{unihall}
	based on HOG and HOD features. The code is not available and we,
	therefore, report only a single point on the precision-recall curve.
	
	\item \AlgDPOMRef{}: This method checks for a human presence on the
	ground plane using \emph{bayesian inference}. Before that, the first
	step is to detect the ground plane and remove it from the 3D processed
	point cloud. After the ground plane elimination, all the points that
	remain will be clustered and segmented as possible person
	detections. The algorithm stops when it finishes to group all the
	regions clustered in the whole depth image.
	
	\item \AlgRCNNRef{}: That uses a region proposal based CNN with
	the~\cite{rcnn} architecture. This architecture was originally used
	for region segmentation and classification. In addition to this
	architecture, it uses the region of interest method
	from~\cite{mva2017}.
	
	\item \AlgRGBCNNRef{}: That uses a RGB CNN-based object detector with region of
	interest selection based on~\cite{mva2017} and selective search
	from~\cite{selective}.
	
	\item \AlgRGBCECDCNNRef{}: That uses a RGB and Depth combined CNN-based object
	detector with CECD channel encoding based on~\cite{mva2017} proposal.
	
\end{itemize}

In addition to these methods, and to allow for additional
experimentation on our \DBGESDPD{} dataset, we have also used the YOLO
(\emph{You Only Look Once}) object detector~\cite{yolov3} to be applied
in the person detection task using depth and RGB data. Our use of the
YOLO strategy comprises two approaches, using the YOLO system as is, and
adapting it to more properly handle the depth information. These are the
two developed systems on this line:

\begin{itemize}
	\item \AlgYOLOvthreeRef{}: RGB object detector~\cite{yolov3} based on
	CNN and bounding box based approximation. This implementation was
	based on the original Yolo V3 architecture and trained with the COCO
	2017 dataset~\cite{coco}. The parameters used to configure the
	architecture were an input image size of $416\times416$, 9 anchors,
	and all the COCO classes.
	
	\item \AlgYOLODepth{}: Depth object detector based on the~\cite{yolov3}
	structure, but modified and retrained to only use depth
	information. This implementation was based on a modified Yolo V3
	architecture to use depth images as input instead of RGB images. This
	implementation is trained with the depth datasets used in this
	paper. The parameters used to configure the architecture, were an
	input image size of 416x416, 9 anchors, and only with the person
	class. No structural changes have been made to the architecture as
	compared to the original version.
	
\end{itemize}
\subsubsection{Evaluation Metrics}
\label{sec:evaluation-metrics}

To provide a detailed view of the performance on the evaluated
algorithms, we have calculated the main standard metrics in a detection
problem, namely $\Precision{}$, $\Recall{}$, and $\Fonescore{}$. The
results are shown both in tables (to provide the precise values),
and in bar graphs (to provide an easier visual comparison).

In the scoring process, the following convention has been adopted
regarding occlusions: In the case that an occluded person is not
detected, it does not generate a detection error if the heads of the
users are occluded in a percentage higher than 50\%. Therefore the
occlusion limitation does not practically affect occlusions between the
bodies of the users, but only considering the occlusion of the upper
body part.

In the results tables, we also include confidence intervals for the
$\Fonescore{}$ metric, for a confidence value of $95\%$, to assess the
statistical significance of the results when comparing different
strategies.  Additionally, the confidence intervals for the
$\Precision{}$, $\Recall{}$, and $\Fonescore{}$ metrics are also shown
in the bar graphs.

In the evaluation with real data, $\Precision{}$-$\Recall{}$ curves are
also included. In the case of the \AlgDPOM{}, \AlgACF{},
\AlgKINECTtwo{}, \AlgUNIHALL, \AlgPCLMunaro{}, \AlgRGBCNN{}, and
\AlgRGBCECDCNN{} algorithms, we use those already provided
in~\cite{mva2017}
and~\cite{Bagautdinov_CVPR_2015_dpom}. For the \AlgYOLODepth{} and
\AlgYOLOvthreeRef{} algorithms, we generate the curves of
$\Precision{}$-$\Recall{}$ through a sweep of the algorithm confidence
threshold. However, building these curves in our proposal is somehow
artificial, leading to curves with strange appearances, as the threshold
sweep is done on the Gaussian distribution threshold. In
Section~\ref{sec:roc-curv-gener} we describe the
$\Precision{}$-$\Recall{}$ curve generation procedure for our proposal,
and the considerations that should be taken into account when addressing
the interpretation of the $\Precision{}$-$\Recall{}$ curves for our
system.

In the case of the $\Fonescore{}$, the obtained curves are scanned
with each of the possible thresholds, to obtain their corresponding
$\Fonescore{}$ and to be able to create the $\Fonescore{}$ vs threshold
curve, which allows choosing the best point or range of points of
the work following the criterion of the best possible $\Fonescore{}$.

An additional parameter of the evaluation metrics is the region on which
they are calculated, including restrictions on the image plane and the
depth range. In some cases (that will be clearly stated when showing the
results), the evaluation metrics have been calculated considering:

\begin{itemize}
	\item An image plane region which is smaller than the full-frame
	size. The objective is to only focus on \emph{full detections} of
	people, avoiding the cases of \emph{incomplete persons} in which they
	could otherwise be partially occluded by the image borders.
	\item A depth range that is smaller than the full depth range of the
	sensor. The objective is avoiding measurements greatly contaminated
	with noise that, for some recording conditions, can be found near the
	image borders or at depths near the sensor sensing limits. 
\end{itemize}

\subsubsection{$\Precision{}$-$\Recall{}$ curves generation for the
	\AlgDPDNetvtwo{} algorithm}
\label{sec:roc-curv-gener}


In this section we provide details on how the threshold to be optimised
in \AlgDPDNetvtwo{} works. The need for this explanation is given
because it is not a conventional confidence threshold like the one that
can be found in algorithms such as YoloV3 (in which the threshold sweep
generates a smooth variation in the performance curves), so that it can
produce effects that are radically different from the usual ones, which
can be reflected in the $\Precision{}$-$\Recall{}$ curves.

\begin{figure*}[!htbp]
	\centering
	\includegraphics[width=0.7\textwidth]{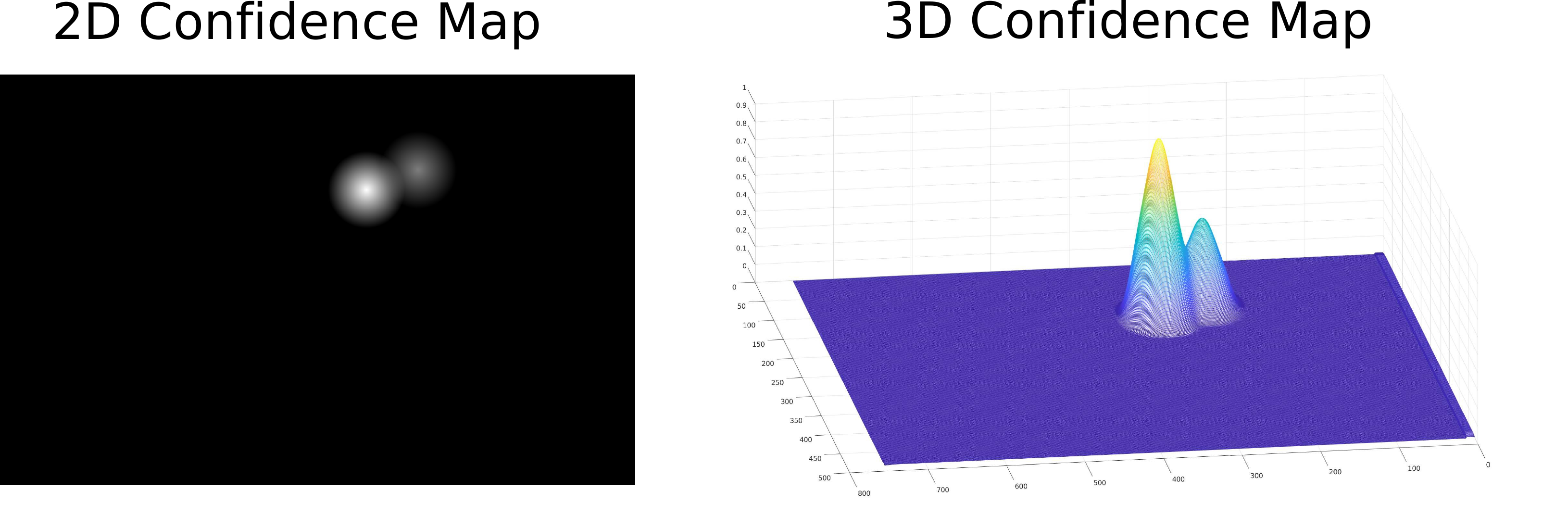}
	\caption{Schematic representation of the 2D confidence map in 3D.}
	\label{fig:gaussiana}
\end{figure*}

Figure~\ref{fig:gaussiana} shows a schematic example of a gaussian-like
confidence map, which could be generated by our system. In the 2D
representation of the figure (left image), we can see two small
gaussian-like regions with a small overlap between them. In the
three-dimensional representation (right image) the apparent overlap
between the two gaussian structures is greater than the one observed in
3D.

The threshold defined in the \AlgDPDNetvtwo{} algorithm would be shown
geometrically as a plane parallel to the
XY plane that would cut the Gaussian distributions at a particular
height, separating them both by their maximum height and by their
overlap at the threshold level. So unlike
conventional thresholds, this threshold will model both the
intergaussian overlap and the maximum confidence peak.

\begin{figure*}[!htbp]
	\centering
	\includegraphics[width=0.7\textwidth]{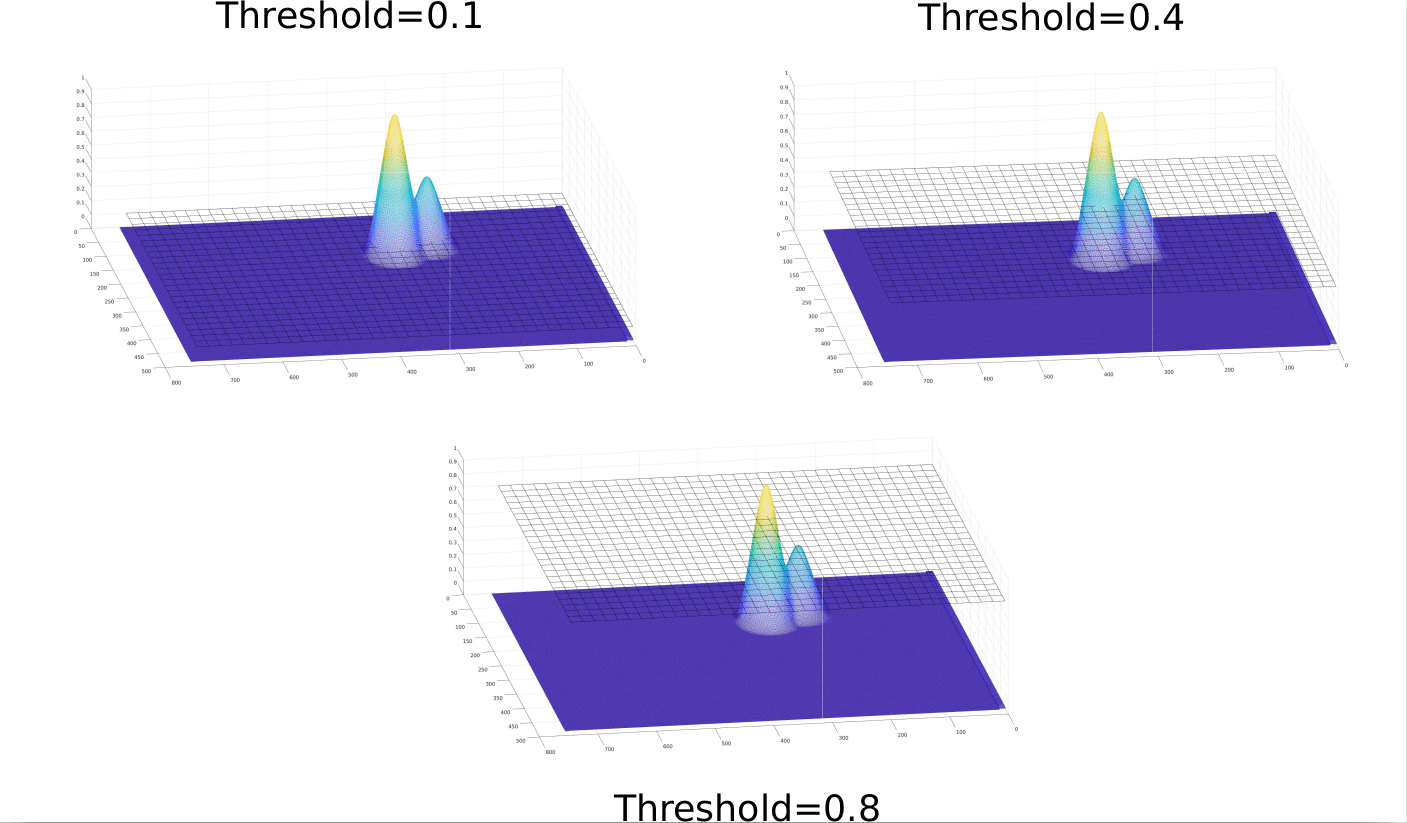}
	\caption{Representation of the effect of different threshold values.}
	\label{fig:gaussiana2}
\end{figure*}

Figure~\ref{fig:gaussiana2} shows the three-dimensional representation
of the two detected gaussians after the application of three thresholds
values at $0.1$, $0.4$, and $0.8$.

In the case of the $0.1$ threshold, it can be seen that the two gaussians
are overlapping by their intersection at the threshold level, so that in
terms of detections, it would generate a single detected person. In the
case of the $0.4$ threshold, it can be observed that the gaussians no
longer have overlap considering the threshold level, so that the two
gaussian centroids are correctly detected. Finally, in the case of
the $0.8$ threshold, only the gaussian with the highest height would
found, since the other one is below the detection threshold, and would
be discarded.

\begin{figure*}[!htbp]
	\centering
	\includegraphics[width=0.7\textwidth]{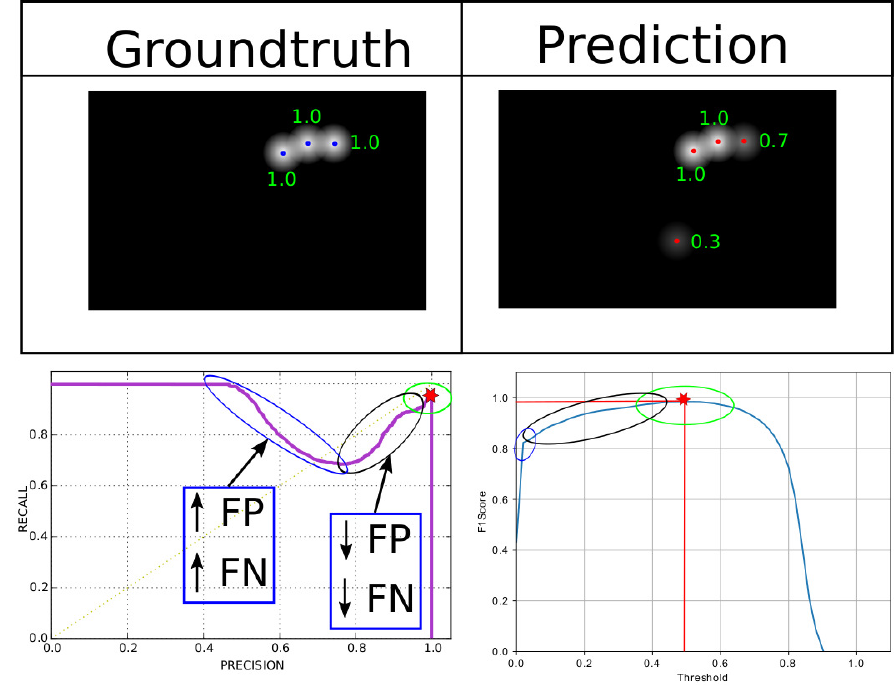}
	\caption{Representation of the threshold variation effect on a
		\DBGFPD{} example. From this example, we
		can conclude that a threshold variation will have a simultaneous
		impact on both the false positive rate (all the low confidence
		gaussians for low threshold levels) and the false negatives (all the
		overlapping gaussians for low threshold levels).}
	\label{fig:gaussiana3}
\end{figure*}

To show a real example on how we build the $\Precision{}$-$\Recall{}$
curves for the \AlgDPDNetvtwo{} system, Figure~\ref{fig:gaussiana3}
includes data obtained for a sample frame of the \DBGFPD{} dataset. The
upper part of the figure includes the ground truth confidence map (left
image), showing three users with gaussians normalized at their maximum
peaks of $1.0$, and the network prediction (right image), with four
estimated gaussians whose maximum peaks correspond to $1.0$, $0.7$ and
$0.3$ (not all the predictions will reach the $1.0$ level).

In the lower part of Figure~\ref{fig:gaussiana3} we can see two
graphs. The one to the left is the $\Precision{}$-$\Recall{}$ graph on
the test set, and the one to the right is the $\Fonescore{}-Threshold$
graph calculated on the training set. In these two graphs we find three
well-differentiated sections that will allow to get an idea of the
distribution of $\Precision{}$-$\Recall{}$ curve values when varying the
threshold value:

\begin{itemize}
	\item The first region, indicated by a blue ellipse, covers a threshold
	range below $0.1$.  In that range, due to the increased gaussian
	overlapping and the acceptance of low confidence gaussians, the number
	of false positives and false negatives increase.
	\item The second region, indicated by a black ellipse, covers a wide threshold
	range below $0.1$ and $0.4$.  In that range, as we increase the
	threshold value the gaussian overlapping will decrease, and also the
	gaussians with low confidence values will be discarded, thus
	decreasing the number of false positives and false negatives.
	\item The third region, indicated by a green ellipse, covers a wide threshold
	range between $0.4$ and $0.8$, and is where the system performs the
	best, with well and correctly separated gaussians, and good rejection
	of low confidence ones. This behaviour leads to the optimum working
	point shown as a red star. 
\end{itemize}

\subsubsection{Training and threshold selection strategy}
\label{sec:thresh-select-strat}

As discussed above, one of the key issues in the \AlgDPDNetvtwo{}
proposal is the correct selection of the detection threshold used in the
output map $C_{polished}$. This selection is an important point because
this threshold is responsible for deciding which gaussians distributions
will be considered as possible person detections. The network and
threshold training have been done using two different approaches:


\begin{itemize}
	\item \emph{Tuned} (Dataset specific) network training and threshold selection:
	In this scenario, the network has been trained on the training subset
	for each specific dataset, and the corresponding threshold is selected
	as that achieving the best $\Fonescore{}$ on this training
	subset. This way we can evaluate the best possible result with the
	threshold tuned to the conditions of each particular dataset. 
	
	\item \emph{Global} network training and threshold selection: In this
	secenario, the network has been trained on all the available training
	subsets, and the threshold has been selected as that achieving the
	best $\Fonescore{}$ on these training subsets. This way we can
	evaluate a realistic performance using the same single threshold for
	all the datasets.
\end{itemize}

\subsection{Results training with Synthetic Data Only}
\label{sec:results-train-with-synth-data}

Our first evaluation task was devoted to the evaluation of till what
extent the training with simulated data could cope with the variability
found in real datasets. To do so, we first run a set of preliminary
experiments using the \DBGESDPD{} as the training data, and both,
GESDPD and EPFL-Lab for testing.



The first row in Table~\ref{tab:gesdpd-gesdpd+epfllab} shows the results of our
proposal trained with the \DBGESDPD{} when evaluated on an independent
subset of the \DBGESDPD{} dataset. The results indicate that the
simulated training data seems to be valid to achieve reasonably good
results when facing similar simulated conditions, with an overall error
below $6.8\%$ and an $\Fonescore{}$ of $96.5\%$.

\begin{table*}[]
	\caption{Results on the \DBGESDPD{} and \DBEPFLLAB{} datasets, trained on the training
		subset of the \DBGESDPD{}.}
	\label{tab:gesdpd-gesdpd+epfllab}
	\resizebox{\textwidth}{!}{%
		\begin{tabular}{|c|c|c|c|c|c|c|c|c|}
			\hline
			& \#FramesTest & \#PeopleTest       &  $FN$  ($FNR\%$)  & $FP$ ($FPR\%$)   & $Error$     & $Precision$ & $Recall$   & $F_{1score}$       \\ \hline
			\DBGESDPD      & $2200$       & $3176$             & $212$ ($6.68\%$)  &  $3$ ($0.09\%$)  & $6.77\%$    & $99.89\%$   & $93.32\%$  & $96.50\pm0.11\%$ \\ \hline
			\DBEPFLLAB     & $950$        & $1959$             & $474$ ($24.07\%$) &  $3$ ($0.15\%$)  & $24.35\%$   & $99.80\%$   & $75.80\%$  & $86.16\pm1.53\%$ \\ \hline
			
		\end{tabular}
	}
\end{table*}


The second row of Table~\ref{tab:gesdpd-gesdpd+epfllab} shows the
results of our proposal trained with the \DBGESDPD{} and evaluated on a
testing subset of the \DBEPFLLAB{} dataset. In this case, the search
area in the image plane was restricted to $280 \times 150$ and the depth
range considered up to $3.5m$. Our objective was to produce an
environment more controlled in which unnecessary noise is removed from
the image, present for example in the ceiling which occupies 40\%of the
image or on the floor. These results indicate that the simulated
training data is not capable of leading to reliable results when facing
a realistic dataset, with an overall error above $24\%$ and an
$\Fonescore{}$ of $86.16\%$.

These preliminary experiments clearly indicate that training with
simulated data only is still far from allowing us to get good results on
realistic data. This conclusion leads us to develop a training procedure
in two stages: first a full training using the simulated data,
and then use a small subset of each of the evaluated datasets to
fine-tune the pre-trained model.

\subsection{Results on real data}
\label{sec:results-dataset-specific-modelsdiscussion}

In this section, we present the results and discussion when evaluating
our proposal on the realistic datasets described in
Section~\ref{sec:datasets}, and using all the available algorithms in
each case. Table~\ref{tab:all-results-all-metrics-overall} shows the results of
all the evaluated algorithms in all the available datasets. Empty cells
in the table are due to the fact that the corresponding algorithms were
not available, so that we have replicated the results published by the
respective authors in the given datasets. 


\begin{table}[!htbp]
	\centering
	\caption{Performance results on all the avaible datasets comparing the``tuned'' and ``global'' versions for the \AlgDPDNetvtwo{} proposal
		($\PrecisionS=\Precision, \RecallS=\Recall{}$ (best result are
		displayed with green background, and orange background indicates
		results within the best one significant bands). }
	\label{tab:all-results-all-metrics-overall}
	\resizebox{\textwidth}{!}{%
		\begin{tabular}{c|ccc|ccc|ccc|ccc|ccc} 
			& \multicolumn{3}{c|}{\DBGFPD{}}                                               & \multicolumn{3}{c|}{\DBKTP{}}                                                & \multicolumn{3}{c|}{\DBUNIHALL{}}                                           & \multicolumn{3}{c|}{\DBEPFLLAB{}}                            & \multicolumn{3}{c}{\DBEPFLCORRIDOR{}}                                         \\\hline                                   
			& $\PrecisionS{}$        & $\RecallS{}$            & $\Fonescore{}$               & $\PrecisionS{}$        & $\RecallS{}$            & $\Fonescore{}$               & $\PrecisionS{}$        & $\RecallS{}$            & $\Fonescore{}$              & $\PrecisionS{}$        & $\RecallS{}$          & $\Fonescore{}$ & $\PrecisionS{}$        & $\RecallS{}$          & $\Fonescore{}$           \\\hline                                  
			\AlgDPDNetvtwo{} tuned   & $99.7$ & \cellcolor{222}$96.36$ & \cellcolor{222}$98.0\pm0.20$ & $96.2$ & $96.3$ & $96.3\pm0.64$                                               & \cellcolor{222}$92.5$ & \cellcolor{222}$99.2$ & \cellcolor{222}$95.7\pm1.30$ & \cellcolor{white}$98.8$ & \cellcolor{222}$94.5$ & \cellcolor{222}$96.6\pm1.18$ & $90.3$ & \cellcolor{222}$80.1$ & \cellcolor{222}$84.9\pm1.05$ \\\hline     
			
			\AlgDPDNetvtwo{} global   & \cellcolor{222}$100.0$ & \cellcolor{white}$95.9$ & \cellcolor{orange}$97.9\pm0.21$ & $95.4$ & $95.1$ & $95.3\pm0.71$ & $91.2$ & $97.3$ & \cellcolor{orange}$94.2\pm1.51$ & $99.5$ & $92.5$ & \cellcolor{orange}$95.9\pm1.30$ & $90.9$ & $76.1$ & \cellcolor{orange}$82.8\pm1.11$  \\\hline     
			
			\AlgYOLOvthree{} & $82.3$                & $86.4$                 & $84.3\pm0.53$                & \cellcolor{222}$99.1$ & \cellcolor{222}$98.2$ & \cellcolor{222}$98.7\pm0.39$  & $84.5$ & $93.1$ & $88.6\pm2.05$                                              & $93.3$ & $93.0$ & $93.2\pm1.65$                               & $58.6$ & $59.8$ & $59.2\pm1.45$  \\\hline                                  
			\AlgYOLODepth{}     & $79.8$                & $55.1$                 & $65.2\pm0.69$                & $91.1$ & $72.3$ & $80.6\pm1.32$                                               & $63.2$ & $52.4$ & $57.3\pm3.19$                                              & $90.2$ & $89.2$ & $89.7\pm1.98$                               & $78.4$ & $47.9$ & $59.5\pm1.45$  \\\hline                                  
			\AlgPCLMunaro{}  &                       &                        &                              & $98.7$ & $76.4$ & $86.1\pm1.16$                                               & $82.5$ & $78.2$ & $80.3\pm2.56$                                              & $96.9$ & $86.5$ & $91.4\pm1.83$                               & $95.0$ & $56.3$ & $70.7\pm1.34$ \\\hline                                   
			\AlgDPOM{}    	 &                       &                        &                              & $95.3$ & $94.5$ & $94.9\pm0.74$                                               & $90.2$ & $98.1$ & \cellcolor{orange}$94.0\pm1.53$                                              & $98.5$ & $85.4$ & $91.5\pm1.82$                               & \cellcolor{222}$96.3$ & $70.9$ & $81.7\pm1.14$ \\\hline                    
			\AlgACF{}        &                       &                        &                              & $87.4$ & $72.7$ & $79.4\pm1.36$                                               & $90.2$ & $85.4$ & $87.7\pm2.11$                                              & $83.8$ & $86.4$ & $85.1\pm2.33$                               & $66.3$ & $40.3$ & $50.1\pm1.47$  \\\hline                                  
			
			\AlgRCNN{}       &                       &                        &                              &        &        &                                                             & $50.1$ & $51.4$ & $50.7\pm3.22$                                              &        &        &                                             &        &        &                \\\hline                                  
			\AlgRGBCNN{}     &                       &                        &                              &        &        &                                                             & $49.7$ & $51.2$ & $50.4\pm3.22$                                              &        &        &                                             &        &        &                \\\hline                                  
			\AlgRGBCECDCNN{} &                       &                        &                              &        &        &                                                             & $52.3$ & $52.3$ & $52.3\pm3.22$                                              &        &        &                                             &        &        &                \\\hline                                  
			\AlgUNIHALL{}    &                       &                        &                              &        &        &                                                             & $86.3$ & $84.5$ & $85.4\pm2.28$                                              &        &        &                                             &        &        &                \\\hline                                  
			
			\AlgKINECTtwo{}  &                       &                        &                              &        &        &                                                             &        &        &                                                            & \cellcolor{222}$99.8$ & $38.2$ & $55.3\pm3.24$                 & $86.3$ & $41.2$ & $55.8\pm1.46$ \\\hline
			
		\end{tabular}
	}
\end{table}

For the \AlgDPDNetvtwo{} algorithm, two results are provided for the
conditions discussed in Section~\ref{sec:thresh-select-strat}: one with
the \emph{tuned} version of the network model and threshold, and the
other with their \emph{global}ly trained versions. The results show, as
expected, a slight decrease in performance when using the global
threshold as compared to the tuned one, but these differences are not
statistically significant. This support the conclusion that the
\AlgDPDNetvtwo{} strategy is robust enough to face very different
training and testing conditions.

In the next subsections we discuss the results for each specific
dataset, providing a graphical comparison of the evaluated metrics,
along with the $\Precision{}$-$\Recall{}$ and $\Fonescore{}-threshold$
curves for the \emph{tuned} approach. The discussion of these curves for
the \emph{global approach} is later addressed in
Section~\ref{sec:disc-prec-recall}.

\subsubsection{Results for the \DBGFPD{} database}
\label{sec:results-dbgfpd-data}

\begin{figure}[!htbp]
	\centering
	\includegraphics[]{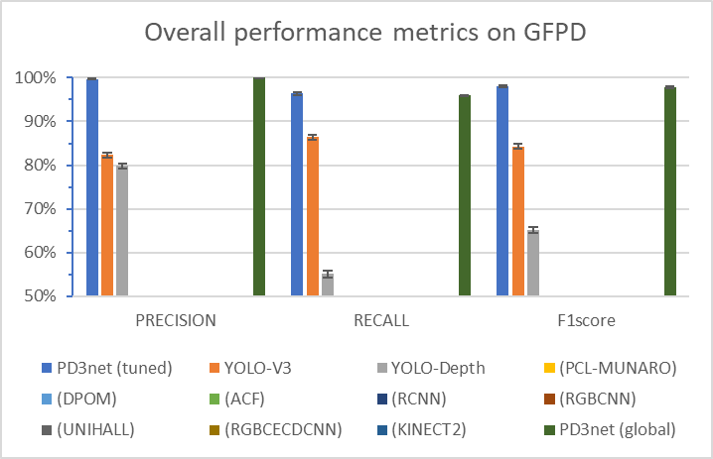}
	\caption{Results on the \DBGFPD{} dataset.}
	\label{fig:results-gfpd}
\end{figure}

The second column of Table~\ref{tab:all-results-all-metrics-overall} and
Figure~\ref{fig:results-gfpd} show the results when evaluating on the
\DBGFPD{} dataset, using our proposal \AlgDPDNetvtwo{} and the
\AlgYOLOvthree{} and \AlgYOLODepth{} algorithms. We could not apply any
of the other proposals described in Section~\ref{sec:experimental-setup}
as they were not readily available.

The results in the table show that both approaches of the
\AlgDPDNetvtwo{} algorithm clearly outperform the YOLO-based
strategies, being the best in the three metrics used and in a
statistically significant way, placing a great distance between it and
\AlgYOLOvthree{} as the second-best solution. The worse results of the
YOLO-based algorithms probably rely in the fact that they were not
designed to deal with depth data, and to the great number of occlusions
and complex situations that \DBGFPD{} contains.

\begin{figure}[!htbp]
	\begin{center}
		\includegraphics[width=0.8\textwidth]{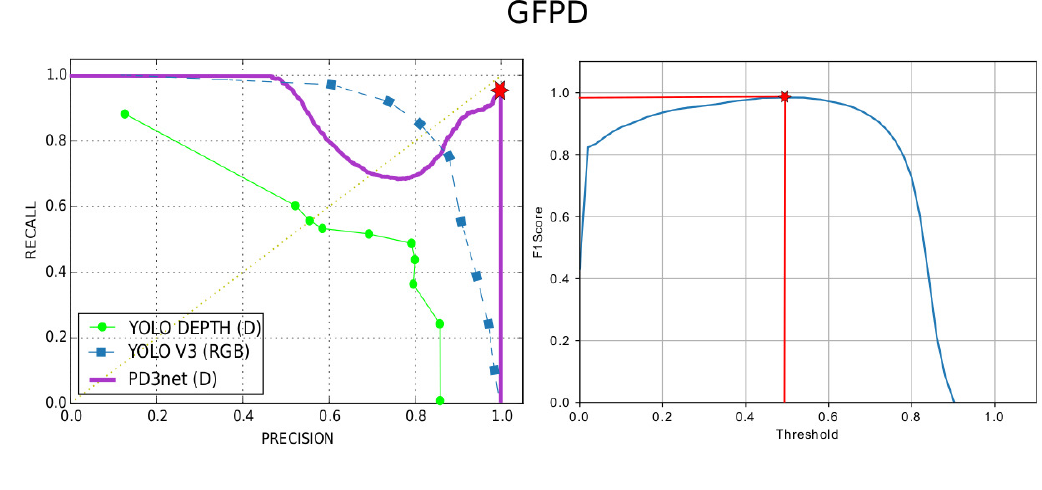}
	\end{center}
	\caption{$\Precision{}$-$\Recall{}$ curve comparison and $\Fonescore{}$-Threshold results for
		the experiments on the \DBGFPD{} dataset.}
	\label{fig:GFPD_PR_F1}
\end{figure}

Figure~\ref{fig:GFPD_PR_F1} shows the $\Precision{}$-$\Recall{}$ curve
corresponding to the behavior of the three algorithms and the
$\Fonescore{}-threshold$ curve corresponding to the \AlgDPDNetvtwo{}
algorithm. As described in Section~\ref{sec:roc-curv-gener}, the
appearance of the curve for our proposal is far from standard, and it is
due to the effect of the threshold sweep procedure so that the area
under the curve should not be taken into account as the comparison
metric, but the actual working point of the algorithm (marked with a red
start). The figure clearly shows that our working point greatly
outperforms the other proposals.

With respect to this $\Precision{}$-$\Recall{}$ curve, $\Fonescore{}$
metric is represented as a function of the threshold sweep used to
generate the $\Precision{}$-$\Recall{}$ curve. This curve shows how the
working point is located in the middle of a reasonably wide and flat
area which outperforms the other two algorithms, indicating that its
sensitivity to the threshold value is reduced.

The capabilities of our proposal are further established in the next
sections where the availability of performance metrics on additional
datasets and with a broad range of different algorithms are exploited in
the comparisons.

\subsubsection{Results for the \DBKTP{} database}
\label{sec:results-dbktp-data}

\begin{figure}[!htbp]
	\centering
	\includegraphics[]{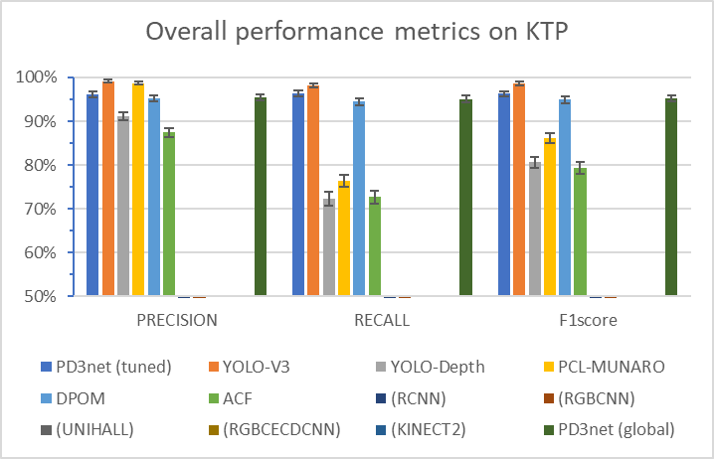}
	\caption{Results on the \DBKTP{} dataset.}
	\label{fig:results-ktp}
\end{figure}

The third column of Table~\ref{tab:all-results-all-metrics-overall} and
Figure~\ref{fig:results-ktp} show the results when evaluating on the
\DBKTP{} dataset, using our proposal \AlgDPDNetvtwo{}, the
\AlgYOLOvthree{} and \AlgYOLODepth{} algorithms, and three other
proposals from the literature (\AlgDPOM{}, \AlgACF{} and
\AlgPCLMunaro{}). 

From the table, it is surprising the top performance of
the \AlgYOLOvthree{} algorithm in terms of $\Fonescore{}$, with
statistically significant different as compared with the second-best
result (achieved by our \AlgDPDNetvtwo{}). In this case, the
\AlgYOLOvthree{} system obtains better results than all the other
proposals as the KTP database does not contain a lot of hard occlusions
and has been prepared from a low frontal perspective, which are the
perfect conditions for the operation of \AlgYOLOvthree{}, whose training
images are also low frontal and with almost no occlusions.

\begin{figure}[!htbp]
	\begin{center}
		\includegraphics[width=\textwidth]{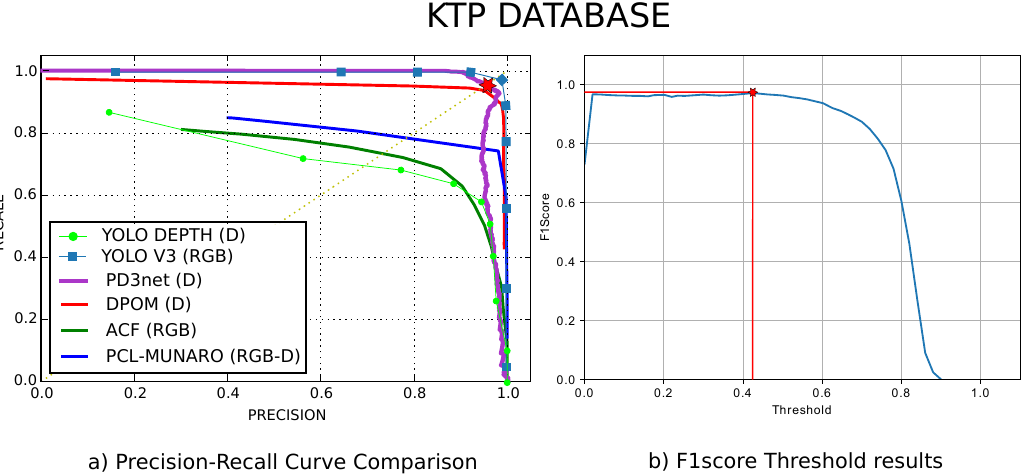}
	\end{center}
	\caption{$\Precision{}$-$\Recall{}$ curve comparison and $\Fonescore{}$-Threshold results for
		the experiments on the \DBKTP{} dataset.}
	\label{fig:KTP_PR_F1}
\end{figure}

Figure~\ref{fig:KTP_PR_F1} shows the $\Precision{}$-$\Recall{}$ curve
corresponding to the behavior of the five algorithms and the
$\Fonescore{}-Threshold$ curve corresponding to the \AlgDPDNetvtwo{}
algorithm. Again, the later curve shows how the working point exhibits
a wide flat area, indicating that its sensitivity to the threshold value
is small. In addition to this, we can see that \AlgYOLOvthree{} shows a
$\Precision{}$-$\Recall{}$ curve that is near the perfection in
\DBKTP{}, compared to \AlgDPOM{} or \AlgDPDNetvtwo{}.


\subsubsection{Results for the \DBUNIHALL{} database}
\label{sec:results-unihall-data}

\begin{figure}[!htbp]
	\centering
	\includegraphics[]{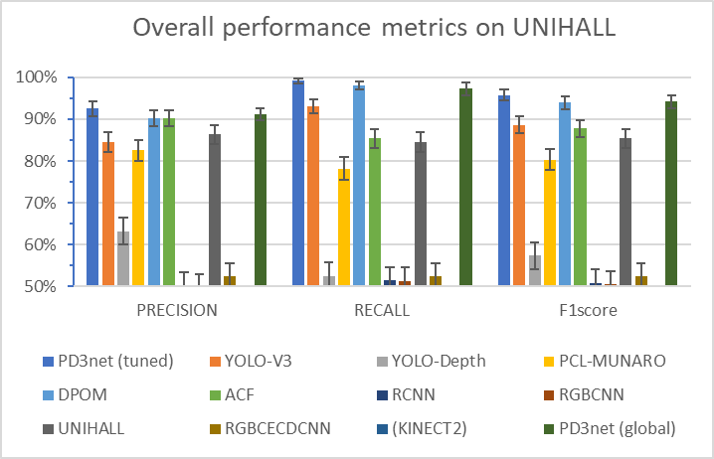}
	\caption{Results on the \DBUNIHALL{} dataset.}
	\label{fig:results-unihall}
\end{figure}

The fourth column of Table~\ref{tab:all-results-all-metrics-overall} and
Figure~\ref{fig:results-unihall} show the results when evaluating on the
\DBKTP{} dataset, using our proposal \AlgDPDNetvtwo{}, the
\AlgYOLOvthree{} and \AlgYOLODepth{} algorithms, and three other
proposals from the literature (\AlgDPOM{}, \AlgACF{}, \AlgPCLMunaro{}
and \AlgKINECTtwo{}). 

In this case, the \AlgDPDNetvtwo{} algorithm is the best one in terms of
the three evaluated metrics, but its improvement as compared with the
second best (\AlgDPOM{}) is not statistically significant.

\begin{figure}[!htbp]
	\begin{center}
		\includegraphics[width=\textwidth]{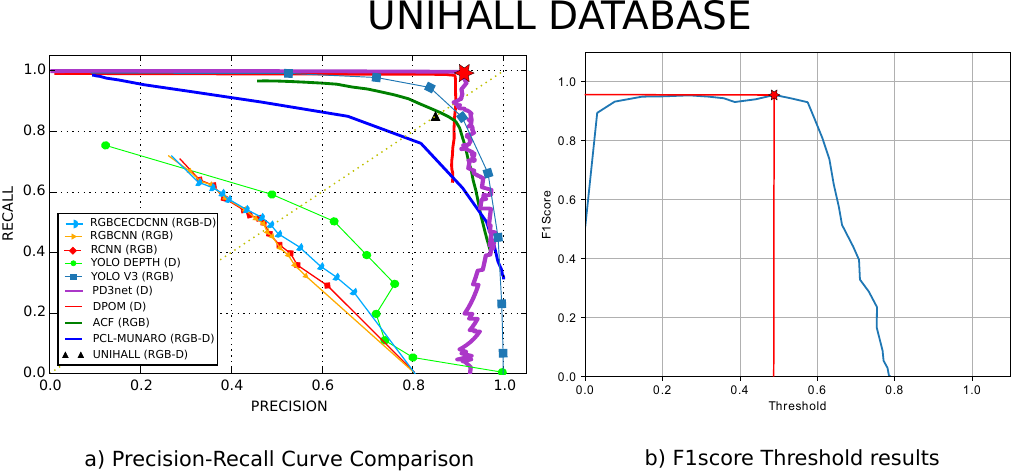}
	\end{center}
	\caption{$\Precision{}$-$\Recall{}$ curve comparison and $\Fonescore{}$-Threshold results for
		the experiments on the \DBUNIHALL{} dataset.}
	\label{fig:UNIHALL_PR_F1}
\end{figure}

Figure~\ref{fig:UNIHALL_PR_F1} shows the $\Precision{}$-$\Recall{}$
curve and the $\Fonescore{}$ sweep curve corresponding to
\AlgDPDNetvtwo{} algorithm. With respect to the $\Fonescore{}$ metric,
the curve shows a stable working point that falls sharply compared to
other databases from a threshold of $0.6$, so we can see that the
working point represented by a sharp peak in the
$\Precision{}$-$\Recall{}$ curve is, in fact, a long-range of stable
working points. In this case, the $\Precision{}$-$\Recall{}$ curve
clearly shows that the proposed algorithm (\AlgDPDNetvtwo{}) has the
best possible working point, very closely followed as the second-best
solution by \AlgDPOM{} and with a large distance in this case to the
best third solution represented by \AlgYOLOvthree{}, which is again
affected by the presence of strong occlusions in this dataset.

\subsubsection{Results for the \DBEPFLLAB{} database}
\label{sec:results-epfllab-data}

\begin{figure}[!htbp]
	\centering
	\includegraphics[]{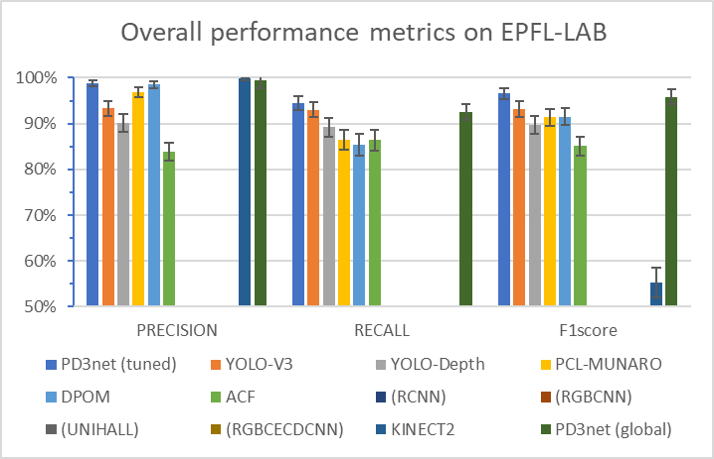}
	\caption{Results on the \DBEPFLLAB{} dataset.}
	\label{fig:results-epfllab}
\end{figure}

The fifth column of Table~\ref{tab:all-results-all-metrics-overall} and
Figure~\ref{fig:results-epfllab} show the results when evaluating on the
\DBEPFLLAB{} dataset, using our proposal \AlgDPDNetvtwo{}, the
\AlgYOLOvthree{} and \AlgYOLODepth{} algorithms, and seven other
proposals from the literature (\AlgDPOM{}, \AlgACF{}, \AlgPCLMunaro{},
\AlgRCNN{}, \AlgRGBCNN{}, \AlgRGBCECDCNN{}, and \AlgUNIHALL{}). 

In this case, the \AlgDPDNetvtwo{} algorithm is the best in terms of the
three evaluated metrics, and its improvements as compared with the
second best (\AlgYOLOvthree{}) are again, statistically significant.

\begin{figure}[!htbp]
	\begin{center}
		\includegraphics[width=\textwidth]{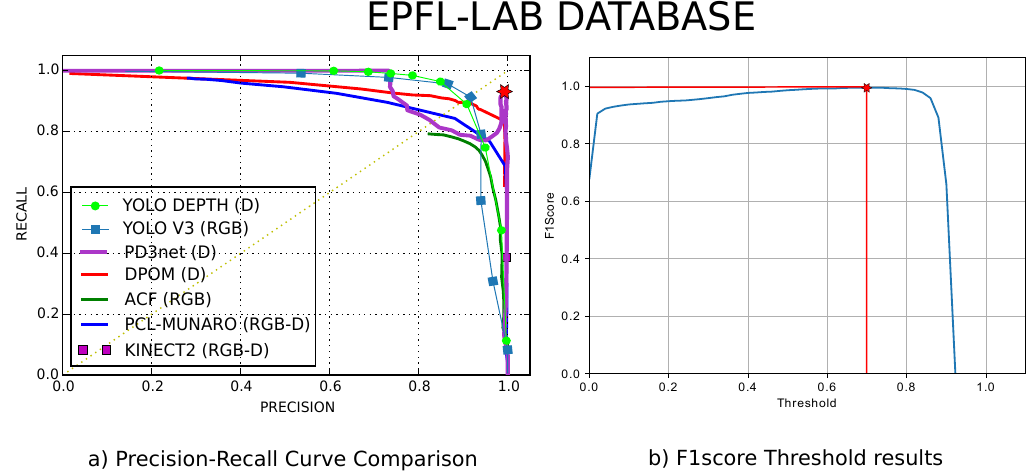}
	\end{center}
	\caption{$\Precision{}$-$\Recall{}$ curve comparison and $\Fonescore{}$-Threshold results for
		the experiments on the \DBEPFLLAB{} dataset.}
	\label{fig:EPFLLAB_PR_F1}
\end{figure}

Figure~\ref{fig:EPFLLAB_PR_F1} shows the $\Precision{}$-$\Recall{}$
curve and the $\Fonescore{}$ sweep curve corresponding to
\AlgDPDNetvtwo{} algorithm. Again, the best algorithm in terms of
working point in the $\Precision{}$-$\Recall{}$ curve is
\AlgDPDNetvtwo{}. This curve exhibits a very different behavior from the
one seen in previous datasets, forming a very steep slope and peak. To
demonstrate that this peak represents a large number of good and stable
working points, the $\Fonescore{}$ graph clearly shows how there is a
range of threshold values (from $0.6$ to $0.8$ that is practically
stable in terms of $\Fonescore{}$. 

\subsubsection{Results for the \DBEPFLCORRIDOR{} database}
\label{sec:results-epflcorridor-data}

\begin{figure}[!htbp]
	\centering
	\includegraphics[]{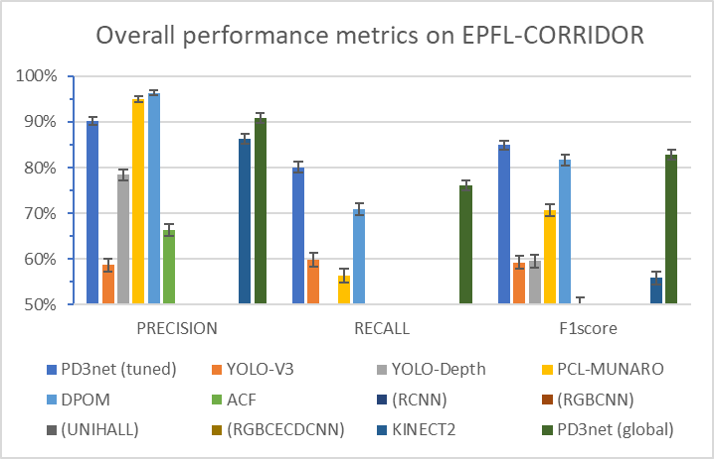}
	\caption{Results on the \DBEPFLCORRIDOR{} dataset.}
	\label{fig:results-epfl-corridor}
\end{figure}

The sixth column of Table~\ref{tab:all-results-all-metrics-overall} and
Figure~\ref{fig:results-epfl-corridor} show the results when evaluating
on the \DBEPFLCORRIDOR{} dataset, using our proposal \AlgDPDNetvtwo{},
the \AlgYOLOvthree{} and \AlgYOLODepth{} algorithms, and four other
proposals from the literature (\AlgDPOM{}, \AlgACF{}, \AlgPCLMunaro{},
and \AlgKINECTtwoRef{}). In this case, the \AlgDPDNetvtwo{} algorithm is
again the best in terms of recall and $\Fonescore{}$ metrics and the
third in terms of precision (although the good results in \Precision{}
by the \AlgDPOM{} algorithm are related to a poor behaviour in terms of
\Recall). Its improvements in terms of \Recall{} and $\Fonescore{}$
compared to the second best algorithm (\AlgDPOM{}) are statistically
significant.

\begin{figure}[!htbp]
	\begin{center}
		\includegraphics[width=\textwidth]{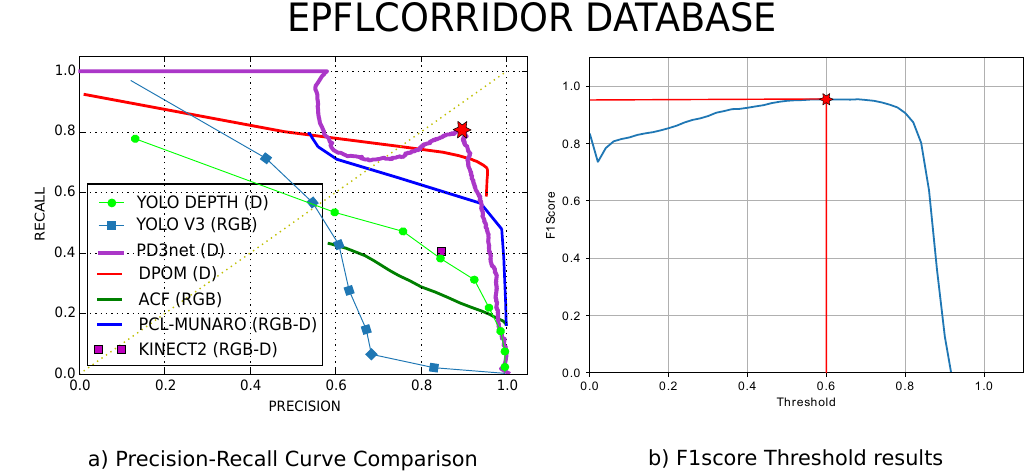}
	\end{center}
	\caption{$\Precision{}$-$\Recall{}$ curve comparison and $\Fonescore{}$-Threshold results for
		the experiments on the \DBEPFLCORRIDOR{} dataset.}
	\label{fig:EPFCORRIDOR_PR_F1}
\end{figure}

Figure~\ref{fig:EPFCORRIDOR_PR_F1} shows the $\Precision{}$-$\Recall{}$
curve and the $\Fonescore{}$ sweep curve corresponding to
\AlgDPDNetvtwo{} algorithm. \DBEPFLCORRIDOR{} is one of the most complex
databases, bringing together difficulties such as many occlusions and
people very close together, detection in small spaces with a lot of
perspectives, great noise in the depth data and many false detections
due to people occluded by the image borders. All these issues lead to
algorithms like \AlgYOLOvthree{} to drastically reduce their
performance, while \AlgDPDNetvtwo{} manages to keep a more robust
performance, this time with a less stable working point in terms of
$\Fonescore{}$, as compared to the results obtained in other
databases. The second best system is again \AlgDPOM{},in this case with
significant differences in the results in terms of working
point. Finally, in third place we find \AlgPCLMunaro{} closely following
the \AlgDPOM{} but with remarkable differences in their
$\Precision{}$-$\Recall{}$ curves.

\subsubsection{Average results for all the available datasets}
\label{sec:average-results-all-datasets-specific}

\begin{figure}[!htbp]
	\centering
	\includegraphics[]{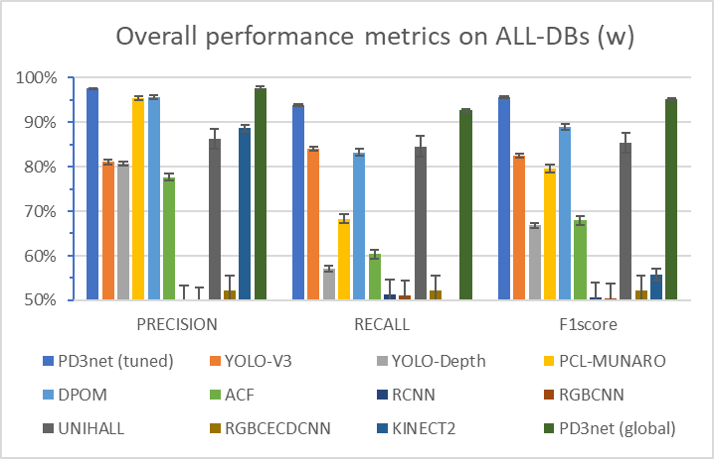}
	\caption{Average weighted results using all the available datasets.}
	\label{fig:results-alldbs-w}
\end{figure}

Table~\ref{tab:results-alldbs-w-overall} and
Figure~\ref{fig:results-alldbs-w} show the weighted average results when
evaluating all the available datasets using all the available
algorithms. They have been calculated integrating the results from the
above sections, weighting each result according to the number of ground
truth elements of each testing subset.

\begin{table}[!htbp]
	\centering
	\caption{Average weighted results using all the available datasets.}
	\label{tab:results-alldbs-w-overall}
	\resizebox{0.55\columnwidth}{!}{%
		\begin{tabular}{ccccccc} 
			& $\Precision{}$ & $\Recall{}$ & $\Fonescore{}$   \\\hline
			\multirow{1}{*}{\AlgDPDNetvtwo{}} tuned  & $97.51$ & \cellcolor{222}$93.80$ & \cellcolor{222}$95.62 \pm 0.24$ \\\hline
			\multirow{1}{*}{\AlgDPDNetvtwo{}} global & \cellcolor{222}$97.68$ & $92.59$ & $95.07 \pm 0.25$ \\\hline
			\AlgYOLOvthree{} & $81.02$ & $84.05$ & $82.51 \pm 0.45$ \\\hline
			\multirow{1}{*}{\AlgYOLODepth{}}     & $80.75$ & $57.08$ & $66.88 \pm 0.55$ \\\hline
			\AlgPCLMunaro{}  & $95.29$ & $68.31$ & $79.57 \pm 0.80$ \\\hline
			\AlgDPOM{}       & $95.57$ & $83.19$ & $88.95 \pm 0.62$ \\\hline
			\AlgACF{}        & $77.67$ & $60.35$ & $67.92 \pm 0.93$ \\\hline
			\AlgRCNN{}       & $50.10$ & $51.40$ & $50.74 \pm 3.22$ \\\hline
			\AlgRGBCNN{}     & $49.70$ & $51.20$ & $50.44 \pm 3.22$ \\\hline
			\AlgUNIHALL{}    & $86.30$ & $84.50$ & $85.39 \pm 2.28$ \\\hline
			\AlgRGBCECDCNN{} & $52.30$ & $52.30$ & $52.30 \pm 3.22$ \\\hline
			\AlgKINECTtwo{}  & $88.58$ & $40.69$ & $55.77 \pm 1.33$ \\\hline
		\end{tabular}
	}
\end{table}

In the average comparison over all the datasets, we can observe that the
statistical significance of the results is increased, provided the
higher number of considered samples. In terms of $\Precision{}$ the best
algorithm is \AlgDPDNetvtwo{}, with \AlgDPOM{} and \AlgPCLMunaro{}
coming second and third. In terms of $\Recall{}$ the differences
increase, with a large distance among the first three best
proposals. The top system is again \AlgDPDNetvtwo{}, followed by
\AlgDPOM{} and \AlgYOLOvthree{}. Finally, considering the $\Fonescore{}$
that relates the two previous metrics offering a final joint one,
\AlgDPDNetvtwo{} comes first, clearly surpassing \AlgDPOM{}, which is
the second-best method with a wide statistical significance.

\subsubsection{Discussion on the $\Precision{}-\Recall{}$ curves for the
	\emph{Global} training approach}
\label{sec:disc-prec-recall}

In this section we provide details on the comparison of our proposal
with the other state of the art methods considering the
$\Precision{}-\Recall{}$ and $\Fonescore{}-Threshold$ curves, when the
network model and the threshold have been trained using all the
available training subsets (referred to as the \emph{Global} approach in
Section~\ref{sec:thresh-select-strat}.

\begin{figure}[!htbp]
	\centering
	\includegraphics[width=\textwidth]{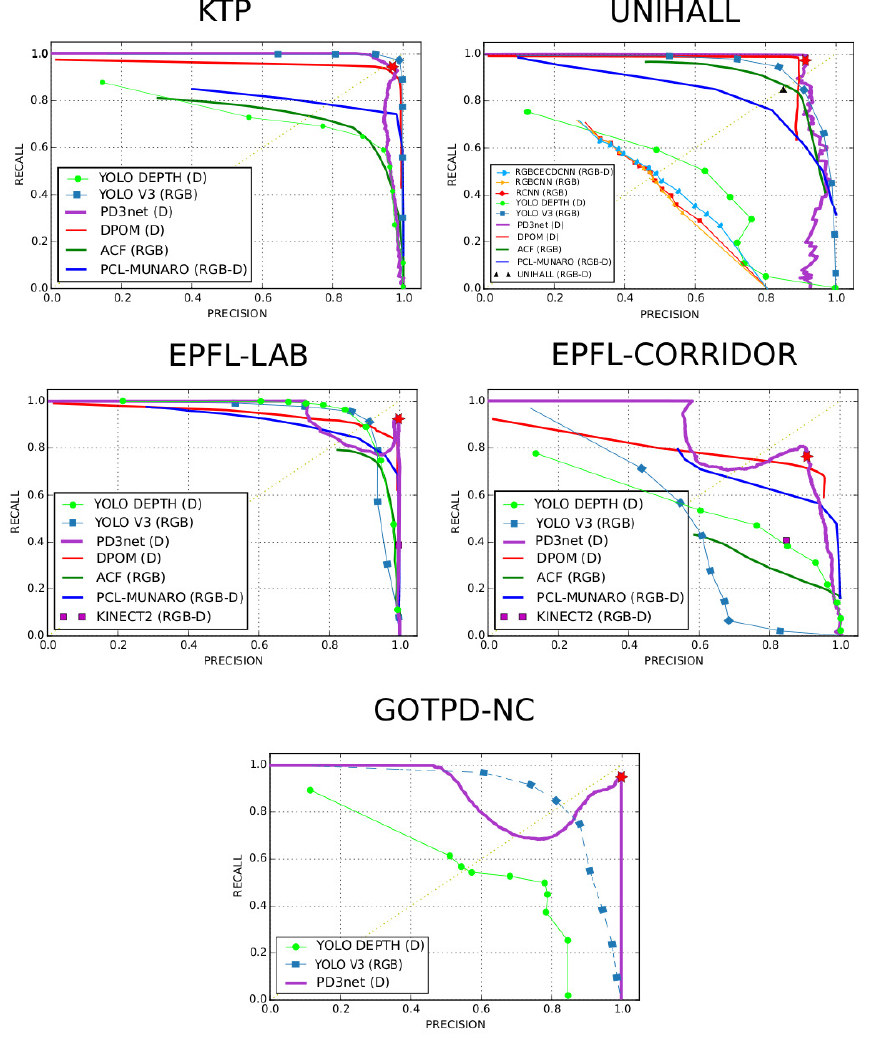}
	\caption{$\Precision{}$-$\Recall{}$ curve comparison using a single global
		network model and a single global threshold for all datasets.}
	\label{fig:overall-roc}
\end{figure}

Figure~\ref{fig:overall-roc} shows the $\Precision$-$\Recall$ curves for
all the datasets and different algorithms.

The curves obtained in this case for each dataset are very similar to
those found in Figures~\ref{fig:GFPD_PR_F1}, \ref{fig:KTP_PR_F1},
\ref{fig:UNIHALL_PR_F1}, \ref{fig:EPFLLAB_PR_F1}
and~\ref{fig:EPFCORRIDOR_PR_F1}, which is consistent with the
performance metric results in
Table~\ref{tab:all-results-all-metrics-overall}, which showed non
significant differences between the \emph{tuned} and \emph{global}
approaches.

\begin{figure}[!htbp]
	\centering
	\includegraphics[width=0.65\columnwidth]{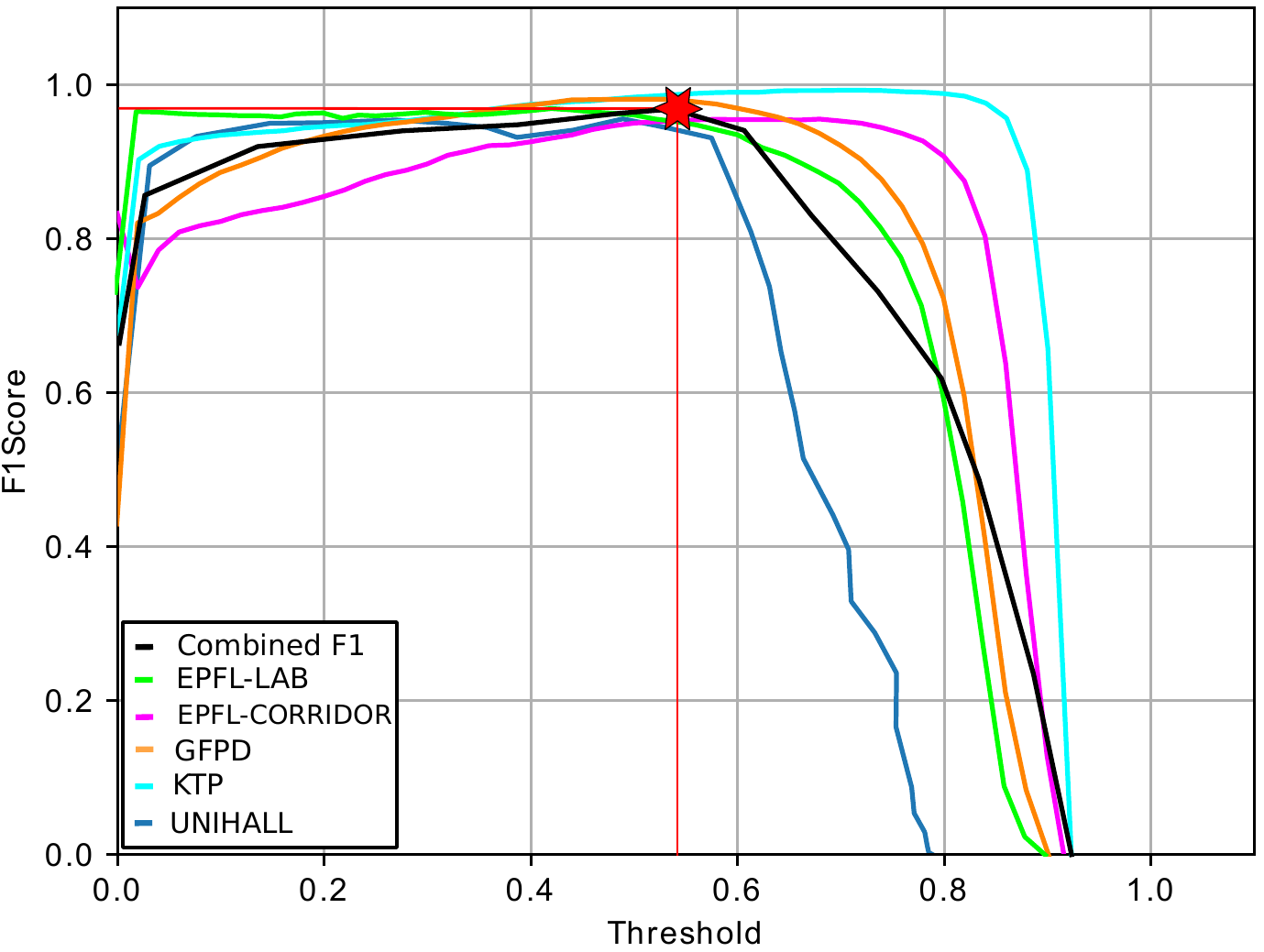}
	\caption{$\Fonescore{}$-Threshold results in the experiments using a
		single global network model and a single global threshold for all
		datasets.}
	\label{fig:overall-roc-variation}
\end{figure}

Figure~\ref{fig:overall-roc-variation} shows the
$\Fonescore{}$-Threshold curve for the proposed \AlgDPDNetvtwo{}
algorithm and the different datasets. We also include the average
curve (labeled ``Combined F1''), from which a single threshold of $0.54$
was selected, as the one with the maximum average $\Fonescore{}$. The
fact that this best threshold is located around the middle of the
threshold span also supports the robustness of the proposed strategy.

\subsection{Computational Performance Evaluation}
\label{sec:perf-eval}

The average frame rate of the system is 42 FPS (frames per second),
benchmarked on a conventional Linux desktop PC, with a Processor
Intel\textregistered Core(TM) i7-6700K CPU @ 4.00 GHz with 64 GB of RAM,
and an NVIDIA GTX-1080 TI GPU.

\section{Conclusions}
\label{sec:conclusions}

This work proposes a new people detection method with depth maps, based
in a non-conventional convolutional neural network approximation. This
approximation was specifically designed to detect people only using the
depth data captured by different types of depth sensor technology(ToF,
active stereo, or structured light).

This article includes a very thorough evaluation and comparison with
different methods of detection of people from the state of the art, both
classical and based on DNN. All this evaluation has been carried out on
5 different RGB-D image databases, each one using a different depth
sensor, all this together with a rigorous experimental evaluation
process. The scenes used for training and those used for evaluation are
as realistic as possible, including a wide variety of users,
backgrounds, and elements in them.

The method proposed has been evaluated with a detection threshold for
each of the databases and with a common threshold for all of them,
making the analysis of the two cases in depth. In addition, an
evaluation has been carried out on all the databases and on each one of
them specifically highlighting the most important results and details.


In each of the databases used our method obtains if not the best, of the
best metrics of all the state of the art systems evaluated, using only
depth maps as input. This system beats in many occasions RGB methods
that have revolutionized the state of the art like \AlgYOLOvthree{} or
classic methods that use very outstanding depth maps like
\AlgDPOM{}. The method used works well even in situations with a large
number of users and occlusions.


The proposed method has worked well on three different depth sensor
technologies, is quite general in that sense. This method requires
training with a synthetically generated database and a small amount of
data captured by the camera to allow it to be coupled with reality, but
does not require a camera calibration process to know the intrinsic and
extrinsic data of the sensor.

\section{Acknowledgments}
\label{sec:acknowledgements}

This work has been partially supported by the Spanish Ministry of Economy and
Competitiveness under projects HEIMDAL-UAH (TIN2016-75982-C2-1-R) and
ARTEMISA (TIN2016-80939-R) and by the University of Alcalá under projects  
ACERCA (CCG2018/EXP-019) and ACUFANO (CCG19/IA-024).

\bibliographystyle{ieee}
\bibliography{DPDnet}

\begin{thebibliography}{10}\itemsep=-1pt

\bibitem{Aguilar2017}
W.~G. Aguilar, M.~A. Luna, J.~F. Moya, V.~Abad, H.~Ruiz, H.~Parra, and
  W.~Lopez.
\newblock Cascade classifiers and saliency maps based people detection.
\newblock In L.~T. De~Paolis, P.~Bourdot, and A.~Mongelli, editors, {\em
  Augmented Reality, Virtual Reality, and Computer Graphics}, pages 501--510,
  Cham, 2017. Springer International Publishing.

\bibitem{segnet}
V.~Badrinarayanan, A.~Kendall, and R.~Cipolla.
\newblock {SegNet}: {A} deep convolutional encoder-decoder architecture for
  image segmentation.
\newblock {\em CoRR}, abs/1511.00561, 2015.

\bibitem{Bagautdinov_CVPR_2015_dpom}
T.~{Bagautdinov}, F.~{Fleuret}, and P.~{Fua}.
\newblock Probability occupancy maps for occluded depth images.
\newblock In {\em 2015 IEEE Conference on Computer Vision and Pattern
  Recognition (CVPR)}, pages 2829--2837, June 2015.

\bibitem{Slawomir2016}
S.~{Bak}, M.~{San Biagio}, R.~{Kumar}, V.~{Murino}, and F.~{Brémond}.
\newblock Exploiting feature correlations by brownian statistics for people
  detection and recognition.
\newblock {\em IEEE Transactions on Systems, Man, and Cybernetics: Systems},
  47(9):2538--2549, Sep. 2017.

\bibitem{blender}
{Blender Online Community}.
\newblock Blender - a {3D} modelling and rendering package, 2018.

\bibitem{bochkovskiy2020yolov4}
A.~Bochkovskiy, C.-Y. Wang, and H.-Y.~M. Liao.
\newblock {YOLOv4}: Optimal speed and accuracy of object detection, 2020.

\bibitem{cao2017}
Y.~Cao, C.~Shen, and H.~T. Shen.
\newblock Exploiting depth from single monocular images for object detection
  and semantic segmentation.
\newblock {\em IEEE Transactions on Image Processing}, 26(2):836--846, 2017.

\bibitem{Chan2008}
A.~Chan, Z.-S. Liang, and N.~Vasconcelos.
\newblock Privacy preserving crowd monitoring: Counting people without people
  models or tracking.
\newblock In {\em Computer Vision and Pattern Recognition, 2008. CVPR 2008.
  IEEE Conference on}, pages 1--7, June 2008.

\bibitem{TsongYi2010}
T.-Y. Chen, C.-H. Chen, D.-J. Wang, and Y.-L. Kuo.
\newblock A people counting system based on face-detection.
\newblock In {\em Genetic and Evolutionary Computing (ICGEC), 2010 Fourth
  International Conference on}, pages 699--702, Dec 2010.

\bibitem{Xception}
F.~Chollet.
\newblock Xception: Deep learning with depthwise separable convolutions.
\newblock {\em CoRR}, abs/1610.02357, 2016.

\bibitem{ByoungKyu2012}
B.-K. Dan, Y.-S. Kim, Suryanto, J.-Y. Jung, and S.-J. Ko.
\newblock Robust people counting system based on sensor fusion.
\newblock {\em Consumer Electronics, IEEE Transactions on}, 58(3):1013--1021,
  August 2012.

\bibitem{delpizzo2016counting}
L.~Del~Pizzo, P.~Foggia, A.~Greco, G.~Percannella, and M.~Vento.
\newblock Counting people by {RGB} or depth overhead cameras.
\newblock {\em Pattern Recognition Letters}, 81(C):41--50, Oct. 2016.

\bibitem{acf}
P.~Dollár, R.~Appel, and W.~Kienzle.
\newblock Crosstalk cascades for frame-rate pedestrian detection.
\newblock pages 645--659, 10 2012.

\bibitem{acf2}
P.~Dollár, Z.~Tu, P.~Perona, and S.~Belongie.
\newblock Integral channel features.
\newblock 01 2009.

\bibitem{du2019spinenet}
X.~Du, T.-Y. Lin, P.~Jin, G.~Ghiasi, M.~Tan, Y.~Cui, Q.~V. Le, and X.~Song.
\newblock {SpineNet}: Learning scale-permuted backbone for recognition and
  localization, 2019.

\bibitem{deepsft}
D.~Fuentes{-}Jim{\'{e}}nez, D.~Casillas{-}P{\'{e}}rez,
  D.~Pizarro{-}P{\'{e}}rez, T.~Collins, and A.~Bartoli.
\newblock Deep shape-from-template: Wide-baseline, dense and fast registration
  and deformable reconstruction from a single image.
\newblock {\em CoRR}, abs/1811.07791, 2018.

\bibitem{GFPD}
D.~Fuentes-Jimenez, C.~L. Gutierrez, J.~M. Guarasa, C.~Luna, and D.~Pizarro.
\newblock Depth person detection database (gfpd-uah).

\bibitem{gesdpd2019}
D.~Fuentes-Jimenez, C.~Losada-Gutierrez, and R.~Martín-Lopez.
\newblock {GESDPD} depth people detection dataset, 2019.

\bibitem{dpdnet}
D.~Fuentes-Jimenez, R.~Martin-Lopez, C.~Losada-Gutierrez, D.~Casillas-Perez,
  J.~Macias-Guarasa, C.~A. Luna, and D.~Pizarro.
\newblock {DPDnet}: A robust people detector using deep learning with an
  overhead depth camera.
\newblock {\em Expert Systems with Applications}, page 113168, 2019.

\bibitem{galvcik2013real}
F.~Gal\u{a}\'{\i}k and R.~Gargal\'{\i}k.
\newblock Real-time depth map based people counting.
\newblock In {\em 15th International Conference on Advanced Concepts for
  Intelligent Vision Systems - Volume 8192}, ACIVS 2013, pages 330--341,
  Berlin, Heidelberg, 2013. Springer-Verlag.

\bibitem{Gavriilidis2018}
A.~Gavriilidis, J.~Velten, S.~Tilgner, and A.~Kummert.
\newblock Machine learning for people detection in guidance functionality of
  enabling health applications by means of cascaded {SVM} classifiers.
\newblock {\em Journal of the Franklin Institute}, 355(4):2009 -- 2021, 2018.
\newblock Special Issue on Recent advances in machine learning for signal
  analysis and processing.

\bibitem{8954436}
G.Ghiasi, T.Lin, and Q.V.Le.
\newblock {NAS-FPN}: Learning scalable feature pyramid architecture for object
  detection.
\newblock In {\em 2019 IEEE/CVF Conference on Computer Vision and Pattern
  Recognition (CVPR)}, pages 7029--7038, 2019.

\bibitem{rcnn}
R.~B. Girshick, J.~Donahue, T.~Darrell, and J.~Malik.
\newblock Rich feature hierarchies for accurate object detection and semantic
  segmentation.
\newblock {\em CoRR}, abs/1311.2524, 2013.

\bibitem{hdm_net}
V.~Golyanik, S.~Shimada, K.~Varanasi, and D.~Stricker.
\newblock {HDM-Net}: Monocular non-rigid {3D} reconstruction with learned
  deformation model.
\newblock {\em CoRR}, abs/1803.10193, 2018.

\bibitem{Deepfake}
D.~Guera and E.~J. Delp.
\newblock Deepfake video detection using recurrent neural networks.
\newblock {\em 2018 15th IEEE International Conference on Advanced Video and
  Signal Based Surveillance (AVSS)}, pages 1--6, 2018.

\bibitem{Hayashi15}
T.~{Hayashi}, M.~{Nishida}, N.~{Kitaoka}, and K.~{Takeda}.
\newblock Daily activity recognition based on dnn using environmental sound and
  acceleration signals.
\newblock In {\em 2015 23rd European Signal Processing Conference (EUSIPCO)},
  pages 2306--2310, 2015.

\bibitem{resnet16}
K.~He, X.~Zhang, S.~Ren, and J.~Sun.
\newblock Deep residual learning for image recognition.
\newblock In {\em 2016 {IEEE} Conference on Computer Vision and Pattern
  Recognition, {CVPR} 2016, Las Vegas, NV, USA, June 27-30, 2016}, pages
  770--778, 2016.

\bibitem{hu2018}
T.~Hu, H.~Zhang, X.~Zhu, J.~Clunis, and G.~Yang.
\newblock Depth sensor based human detection for indoor surveillance.
\newblock {\em Future Generation Computer Systems}, 88:540 -- 551, 2018.

\bibitem{realsense}
Intel.
\newblock Intel realsense {D435} product.
\newblock \url{https://www.intelrealsense.com/depth-camera-d435/}.

\bibitem{ChiYoon2013}
C.~Y. Jeong, S.~Choi, and S.~W. Han.
\newblock A method for counting moving and stationary people by interest point
  classification.
\newblock In {\em Image Processing (ICIP), 2013 20th IEEE International
  Conference on}, pages 4545--4548, Sept 2013.

\bibitem{swats}
N.~S. Keskar and R.~Socher.
\newblock Improving generalization performance by switching from adam to {SGD}.
\newblock {\em CoRR}, abs/1712.07628.

\bibitem{Deepfake2}
H.~Kim, P.~Garrido, A.~Tewari, W.~Xu, J.~Thies, N.~Nie{\ss}ner, P.~P{\'e}rez,
  C.~Richardt, M.~Zollh{\"o}fer, and C.~Theobalt.
\newblock {Deep Video Portraits}.
\newblock {\em ACM Transactions on Graphics 2018 (TOG)}, 2018.

\bibitem{kinect2}
Kinect-SDK.
\newblock Kinect for windows {SDK} 2.0.
\newblock \url{http://www.microsoft.com/en-us/kinectforwindows/.}, 2014.

\bibitem{kingma2014adam}
D.~P. Kingma and J.~Ba.
\newblock Adam: {A} method for stochastic optimization.
\newblock {\em CoRR}, abs/1412.6980, 2014.

\bibitem{imagenet2012}
A.~Krizhevsky, I.~Sutskever, and G.~E. Hinton.
\newblock {ImageNet} classification with deep convolutional neural networks.
\newblock In F.~Pereira, C.~J.~C. Burges, L.~Bottou, and K.~Q. Weinberger,
  editors, {\em Advances in Neural Information Processing Systems 25}, pages
  1097--1105. Curran Associates, Inc., 2012.

\bibitem{lee2013context}
K.-D. Lee, M.~Y. Nam, K.-Y. Chung, Y.-H. Lee, and U.-G. Kang.
\newblock Context and profile based cascade classifier for efficient people
  detection and safety care system.
\newblock {\em Multimedia Tools and Applications}, 63(1):27--44, 2013.

\bibitem{coco}
T.-Y. Lin, M.~Maire, S.~Belongie, L.~Bourdev, R.~Girshick, J.~Hays, P.~Perona,
  D.~Ramanan, C.~L. Zitnick, and P.~Dollár.
\newblock Microsoft {COCO}: Common objects in context, 2014.

\bibitem{liu2015}
J.~Liu, Y.~Liu, G.~Zhang, P.~Zhu, and Y.~Q. Chen.
\newblock Detecting and tracking people in real time with {RGB-D} camera.
\newblock {\em Pattern Recognition Letters}, 53:16 -- 23, 2015.

\bibitem{unihall2}
M.~Luber, L.~Spinello, and K.~O. Arras.
\newblock People tracking in {RGB-D} data with on-line boosted target models.
\newblock In {\em 2011 {IEEE/RSJ} International Conference on Intelligent
  Robots and Systems, {IROS} 2011, San Francisco, CA, USA, September 25-30,
  2011}, pages 3844--3849. {IEEE}, 2011.

\bibitem{luna2017}
C.~A. Luna, C.~Losada-Gutierrez, D.~Fuentes-Jimenez, A.~Fernandez-Rincon,
  M.~Mazo, and J.~Macias-Guarasa.
\newblock Robust people detection using depth information from an overhead
  time-of-flight camera.
\newblock {\em Expert Syst. Appl.}, 71(C):240--256, Apr. 2017.

\bibitem{moro2018}
A.~Moro, J.~Wakabayashi, T.~Toda, and K.~Umeda.
\newblock A framework for human recognition and counting in restricted area for
  video surveillance.
\newblock In {\em Intelligent Environments (Workshops)}, pages 139--148, 2018.

\bibitem{munaro2014}
M.~Munaro and E.~Menegatti.
\newblock Fast {RGB-D} people tracking for service robots.
\newblock {\em Autonomous Robots}, 37, 10 2014.

\bibitem{artifacts}
A.~Odena, V.~Dumoulin, and C.~Olah.
\newblock Deconvolution and checkerboard artifacts.
\newblock {\em Distill}, 1, 10 2016.

\bibitem{Ramanan2007}
D.~Ramanan, D.~A. Forsyth, and A.~Zisserman.
\newblock {Tracking People by Learning Their Appearance}.
\newblock {\em Pattern Analysis and Machine Intelligence, IEEE Transactions
  on}, 29(1):65--81, Nov. 2006.

\bibitem{yolo2016}
J.~Redmon, S.~K. Divvala, R.~B. Girshick, and A.~Farhadi.
\newblock You only look once: Unified, real-time object detection.
\newblock {\em CoRR}, abs/1506.02640, 2015.

\bibitem{yolov3}
J.~Redmon and A.~Farhadi.
\newblock {YOLOv3}: An incremental improvement.
\newblock {\em CoRR}, abs/1804.02767, 2018.

\bibitem{ren2017}
X.~{Ren}, S.~{Du}, and Y.~{Zheng}.
\newblock Parallel {RCNN}: A deep learning method for people detection using
  rgb-d images.
\newblock In {\em 2017 10th International Congress on Image and Signal
  Processing, BioMedical Engineering and Informatics (CISP-BMEI)}, pages 1--6,
  Oct 2017.

\bibitem{erfnet}
E.~{Romera}, J.~M. {Álvarez}, L.~M. {Bergasa}, and R.~{Arroyo}.
\newblock Efficient convnet for real-time semantic segmentation.
\newblock In {\em 2017 IEEE Intelligent Vehicles Symposium (IV)}, pages
  1789--1794, June 2017.

\bibitem{hoo2016}
H.-c. Shin, H.~Roth, M.~Gao, L.~Lu, Z.~Xu, I.~Nogues, J.~Yao, D.~Mollura, and
  R.~Summers.
\newblock Deep convolutional neural networks for computer-aided detection:
  {CNN} architectures, dataset characteristics and transfer learning.
\newblock {\em IEEE Transactions on Medical Imaging}, 35:1285--1298, 02 2016.

\bibitem{kinect21}
J.~Shotton, A.~Fitzgibbon, M.~Cook, T.~Sharp, M.~Finocchio, R.~Moore,
  A.~Kipman, and A.~Blake.
\newblock Real-time human pose recognition in parts from single depth images.
\newblock volume~56, pages 1297--1304, 06 2011.

\bibitem{unihall}
L.~{Spinello} and K.~O. {Arras}.
\newblock People detection in {RGB-D} data.
\newblock In {\em 2011 IEEE/RSJ International Conference on Intelligent Robots
  and Systems}, pages 3838--3843, Sep. 2011.

\bibitem{Stahlschmidt2013}
C.~Stahlschmidt, A.~Gavriilidis, J.~Velten, and A.~Kummert.
\newblock People detection and tracking from a top-view position using a
  time-of-flight camera.
\newblock In A.~Dziech and A.~Czyazwski, editors, {\em Multimedia
  Communications, Services and Security}, volume 368 of {\em Communications in
  Computer and Information Science}, pages 213--223. Springer Berlin
  Heidelberg, 2013.

\bibitem{Stahlschmidt2014}
C.~Stahlschmidt, A.~Gavriilidis, J.~Velten, and A.~Kummert.
\newblock Applications for a people detection and tracking algorithm using a
  time-of-flight camera.
\newblock {\em Multimedia Tools and Applications}, pages 1--18, 2014.

\bibitem{inception}
C.~Szegedy, V.~Vanhoucke, S.~Ioffe, J.~Shlens, and Z.~Wojna.
\newblock Rethinking the inception architecture for computer vision.
\newblock {\em CoRR}, abs/1512.00567, 2015.

\bibitem{du2015}
Y.~{Tian}, P.~{Luo}, X.~{Wang}, and X.~{Tang}.
\newblock Pedestrian detection aided by deep learning semantic tasks.
\newblock In {\em 2015 IEEE Conference on Computer Vision and Pattern
  Recognition (CVPR)}, pages 5079--5087, June 2015.

\bibitem{selective}
J.~R.~R. Uijlings, K.~E.~A. van~de Sande, T.~Gevers, and A.~W.~M. Smeulders.
\newblock Selective search for object recognition.
\newblock {\em International Journal of Computer Vision}, 104(2):154--171, Sep
  2013.

\bibitem{vera2016counting}
P.~Vera, S.~Monjaraz, and J.~Salas.
\newblock Counting pedestrians with a zenithal arrangement of depth cameras.
\newblock {\em Machine Vision and Applications}, 27(2):303--315, Feb 2016.

\bibitem{Verma2019}
N.~K. Verma, R.~Dev, S.~Maurya, N.~K. Dhar, and P.~Agrawal.
\newblock People counting with overhead camera using fuzzy-based detector.
\newblock In N.~K. Verma and A.~K. Ghosh, editors, {\em Computational
  Intelligence: Theories, Applications and Future Directions - Volume I}, pages
  589--601, Singapore, 2019. Springer Singapore.

\bibitem{villamizar2018}
M.~Villamizar, A.~Mart{\'\i}nez-Gonz{\'a}lez, O.~Can{\'e}vet, and J.-M. Odobez.
\newblock Watchnet: Efficient and depth-based network for people detection in
  video surveillance systems.
\newblock In {\em 2018 15th IEEE International Conference on Advanced Video and
  Signal Based Surveillance (AVSS)}, pages 1--6. IEEE, 2018.

\bibitem{wang2017}
C.~{Wang} and Y.~{Zhao}.
\newblock Multi-layer proposal network for people counting in crowded scene.
\newblock In {\em 2017 10th International Conference on Intelligent Computation
  Technology and Automation (ICICTA)}, pages 148--151, Oct 2017.

\bibitem{wang2016human}
S.~Wang, L.~Chen, Z.~Zhou, X.~Sun, and J.~Dong.
\newblock Human fall detection in surveillance video based on {PCANet}.
\newblock {\em Multimedia tools and applications}, 75(19):11603--11613, 2016.

\bibitem{activations}
B.~Xu, N.~Wang, T.~Chen, and M.~Li.
\newblock Empirical evaluation of rectified activations in convolutional
  network.
\newblock {\em CoRR}, abs/1505.00853, 2015.

\bibitem{zhang2015human}
L.~Zhang, X.~Wu, and D.~Luo.
\newblock Human activity recognition with {HMM-DNN} model.
\newblock In {\em 2015 IEEE 14th International Conference on Cognitive
  Informatics \& Cognitive Computing (ICCI* CC)}, pages 192--197. IEEE, 2015.

\bibitem{zhang2016far}
S.~Zhang, R.~Benenson, M.~Omran, J.~Hosang, and B.~Schiele.
\newblock How far are we from solving pedestrian detection?
\newblock In {\em Proceedings of the IEEE Conference on Computer Vision and
  Pattern Recognition}, pages 1259--1267, 2016.

\bibitem{zhang2018}
S.~Zhang, R.~Benenson, M.~Omran, J.~Hosang, and B.~Schiele.
\newblock Towards reaching human performance in pedestrian detection.
\newblock {\em IEEE Transactions on Pattern Analysis and Machine Intelligence},
  40(4):973--986, April 2018.

\bibitem{Zhang2012}
X.~Zhang, J.~Yan, S.~Feng, Z.~Lei, D.~Yi, and S.~Li.
\newblock Water filling: Unsupervised people counting via vertical {K}inect
  sensor.
\newblock In {\em Advanced Video and Signal-Based Surveillance (AVSS), 2012
  IEEE Ninth International Conference on}, pages 215--220, Sept 2012.

\bibitem{zhao2017}
J.~Zhao, G.~Zhang, L.~Tian, and Y.~Q. Chen.
\newblock Real-time human detection with depth camera via a physical
  radius-depth detector and a {CNN} descriptor.
\newblock In {\em 2017 IEEE International Conference on Multimedia and Expo
  (ICME)}, pages 1536--1541, July 2017.

\bibitem{zhou2017}
K.~{Zhou}, A.~{Paiement}, and M.~{Mirmehdi}.
\newblock Detecting humans in {RGB-D} data with {CNNs}.
\newblock In {\em 2017 Fifteenth IAPR International Conference on Machine
  Vision Applications (MVA)}, pages 306--309, May 2017.

\bibitem{mva2017}
K.~{Zhou}, A.~{Paiement}, and M.~{Mirmehdi}.
\newblock Detecting humans in rgb-d data with cnns.
\newblock In {\em 2017 Fifteenth IAPR International Conference on Machine
  Vision Applications (MVA)}, pages 306--309, May 2017.

\bibitem{zhu2013human}
L.~Zhu and K.-H. Wong.
\newblock Human tracking and counting using the kinect range sensor based on
  adaboost and kalman filter.
\newblock In {\em International Symposium on Visual Computing}, pages 582--591.
  Springer, 2013.

\end{thebibliography}

\end{document}